\documentclass[conference]{IEEEtran}

\usepackage{xcolor}

\ifCLASSINFOpdf
   \usepackage[pdftex]{graphicx}
\else
   \usepackage[dvips]{graphicx}
\fi
\ifCLASSOPTIONcompsoc
  \usepackage[caption=false,font=normalsize,labelfont=sf,textfont=sf]{subfig}
\else
  \usepackage[caption=false,font=footnotesize]{subfig}
\fi
\def\BibTeX{{\rm B\kern-.05em{\sc i\kern-.025em b}\kern-.08em
    T\kern-.1667em\lower.7ex\hbox{E}\kern-.125emX}}

\newcounter{doubleblind}
\usepackage{ifthen}

\begin{document}
%
\title{Should You Go Deeper?\\ Optimizing Convolutional Neural Network Architectures without Training}

\ifthenelse{\value{doubleblind}>0}{
\author{
\IEEEauthorblockN{double-blind reviewing policy}
\IEEEauthorblockA{}

}
}{
\author{
\IEEEauthorblockN{Mats L. Richter}
\IEEEauthorblockA{\textit{Osnabrück University} \\ 
Osnabrueck, Germany \\ 
Email: matrichter@uni-osnabrueck.de }

\and

\IEEEauthorblockN{Julius Schöning}
\IEEEauthorblockA{\textit{Osnabrück University of Applied Sciences}\\
Osnabrueck, Germany \\ 
Email: j.schoening@hs-osnabrueck.de}

\and

\IEEEauthorblockN{Anna Wiedenroth}
\IEEEauthorblockA{\textit{Osnabrück University}\\
Osnabrueck, Germany \\ 
Email: awiedenroth@uni-osnabrueck.de}

\and

\IEEEauthorblockN{Ulf Krumnack}
\IEEEauthorblockA{\textit{Osnabrück University}\\ 
Osnabrueck, Germany\\ 
Email: krumnack@uni-osnabrueck.de}

}
}



%


\maketitle

\begin{abstract}
When optimizing convolutional neural networks (CNN) for a specific image-based task, specialists commonly overshoot the number of convolutional layers in their designs. By implication, these CNNs are unnecessarily resource intensive to train and deploy, with diminishing beneficial effects on the predictive performance.

The features a convolutional layer can process are strictly limited by its receptive field.
By layer-wise analyzing the size of the receptive fields, we can reliably predict sequences of layers that will not contribute qualitatively to the test accuracy in the given CNN architecture.
Based on this analysis, we propose design strategies based on a so-called \emph{border layer}. This layer allows to identify unproductive convolutional layers and hence to resolve these inefficiencies, optimize the explainability and the computational performance of CNNs.
Since neither the strategies nor the analysis requires training of the actual model, these insights allow for a very efficient design process of CNN architectures, which might be automated in the future.

\end{abstract}


%

\section{Introduction}
\begin{figure}[ht!]
	\centering
	\label{fig:tails1}
	\includegraphics[width=\columnwidth,trim=0mm 7mm 0mm 2.5mm, clip]{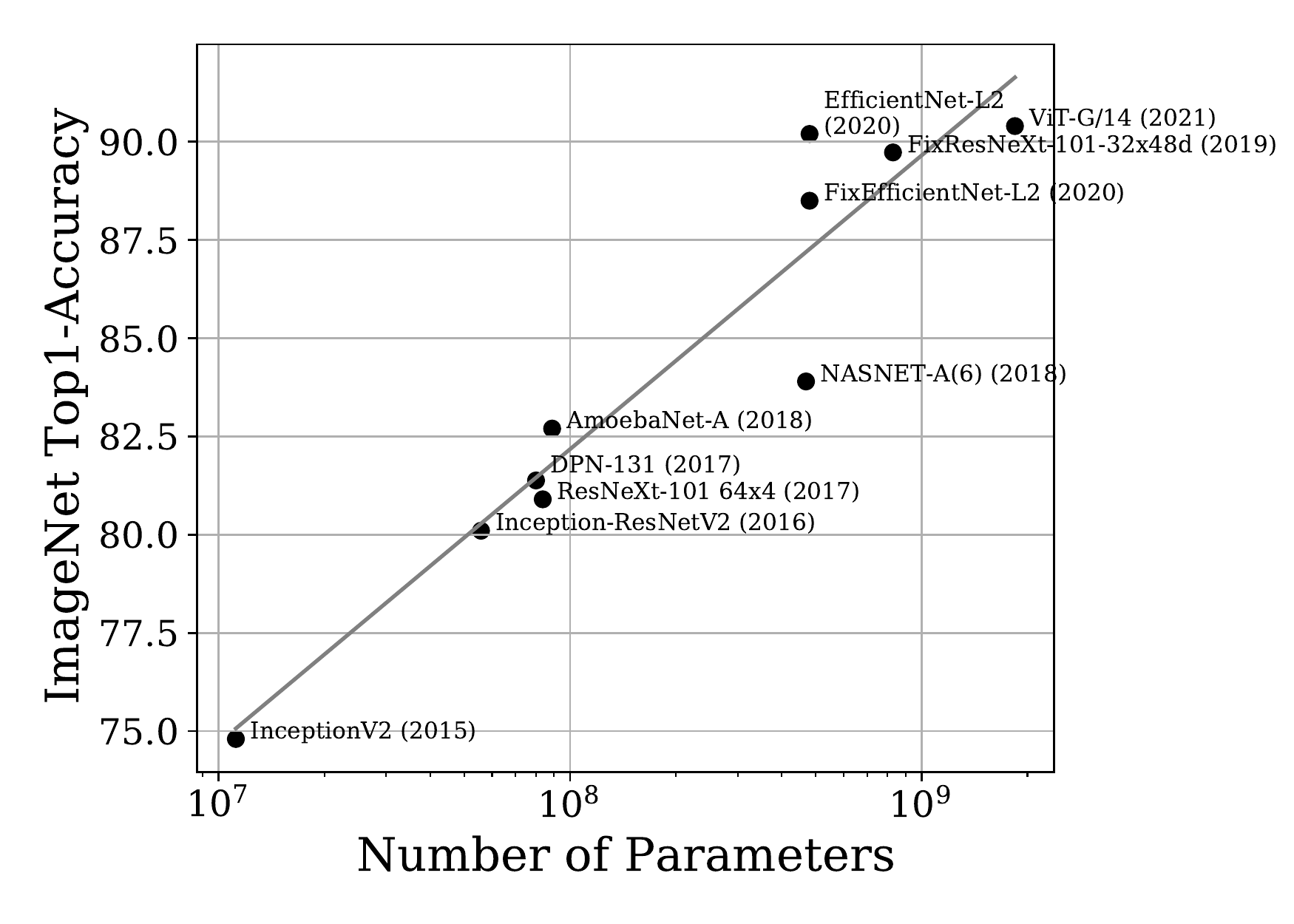}
	\caption{Selected state-of-the-art CNN architectures from recent years show a trend, where improvements in predictive performance strongly coincide with increased number of trainable parameters and computational complexity.}
	\label{fig:sota}
\end{figure}

In recent years, the trend of creating exponentially higher capacity models to gain incremental improvements in predictive performance has dominated the design of convolutional neural network (CNN) classifiers in computer vision.
This is exemplified in Fig.~\ref{fig:sota}, which shows that 92.1\% of the ImageNet-accuracy's variance in state-of-the-art architectures can be explained by an exponential increase in parameters.
Since high-capacity models require also more computational resources to be trained, the architectures also become increasingly uneconomical to train and deploy in practical application scenarios.
One of the most popular design-axis used to increase the model capacity is the number of layers or "depth" of the network \cite{vgg, resnet, inceptionv3}.
Like other parts of the neural architecture design, the optimal depth of a neural architecture for a given problem can currently only be approximated by comparative evaluation of trained models \cite{vgg, resnet, amoebanet, efficientnet}.
This design process is thus firmly founded in trial-and-error.
To move towards a more informed design process that yields also more parameter-efficient models, heuristics based on an understanding of the influence of the neural architecture on the model is needed.
Based on such a heuristic, it should be possible to estimate, ideally before training, whether the addition or removal of layers will have a substantial impact on the predictive performance.

Our approach to such a heuristic for the optimization of CNN architecture depth is based on the interplay between input resolution of the image data and receptive field size of the network's layers.

Our primary contributions can be summarized as answers to the following questions:
\begin{itemize}
    \item Is it possible to predict unproductive subsequences of layers in a neural network?
    Answer: Yes, predicting unproductive sequences for simple sequential CNN architectures and architectures with multiple CNN pathways and skip connections before training is possible by the so-called border layer, cf. Section \ref{sec:seq}.
    \item Since attention mechanisms like SE-Modules 
    \cite{Hu2018}
     effectively provide global context to the feature map, do they change how the inference is distributed?
    Answer: No, the distribution of the inference process is not affected by the use of simple attention mechanisms. Thus, it is still possible to predict unproductive, i.e., unnecessary layers before training, cf. Section \ref{sec:attention}.
    \item Can knowledge about unproductive layers be leveraged as a heuristic to optimize the depth of CNN architectures?
    Answer: Yes, we demonstrate basic strategies that yield reliable improvements in efficiency and predictive performance, cf. Section \ref{sec:design}.
\end{itemize}

\section{Background}\label{sec:Background}
This section discusses the essential background knowledge framing this paper:
First, analysis techniques for judging the quality of intermediate solutions and the processing within hidden layers are introduced.
We further discuss parameter-inefficiencies detectable by these techniques.
Then, the receptive field of convolutional layers is discussed and how its size in sequential and non-sequential architectures is computed.



\subsection{Logistic Regression Probes}
Logistic Regression Probes (LRP) by Alain and Bengio \cite{probes} are a tool for analyzing how the solution quality progresses while the data is propagated through a trained neural network.
The idea is to fit simple regression models on the activation values of each layer. 
By comparing the LRPs' test accuracies with the test accuracy of the full network, it is possible to quantify the extent to which the problem is already solved in each layer.
Typically, LRP accuracies increase layer by layer and approach the model's accuracy, as shown by the red curve in Fig.~\ref{fig:vgg16_cifar10}.
In this paper, the LRP accuracy of a probe computed on the activation values of layer $l$ is referred to as $p_l$.

\subsection{Saturation Values and \textit{Tail Patterns}}\label{sec:tailPatterns}
\begin{figure}[ht]
	\centering
	\includegraphics[width=0.9\columnwidth,trim=0mm 3.5mm 0mm 11mm, clip]{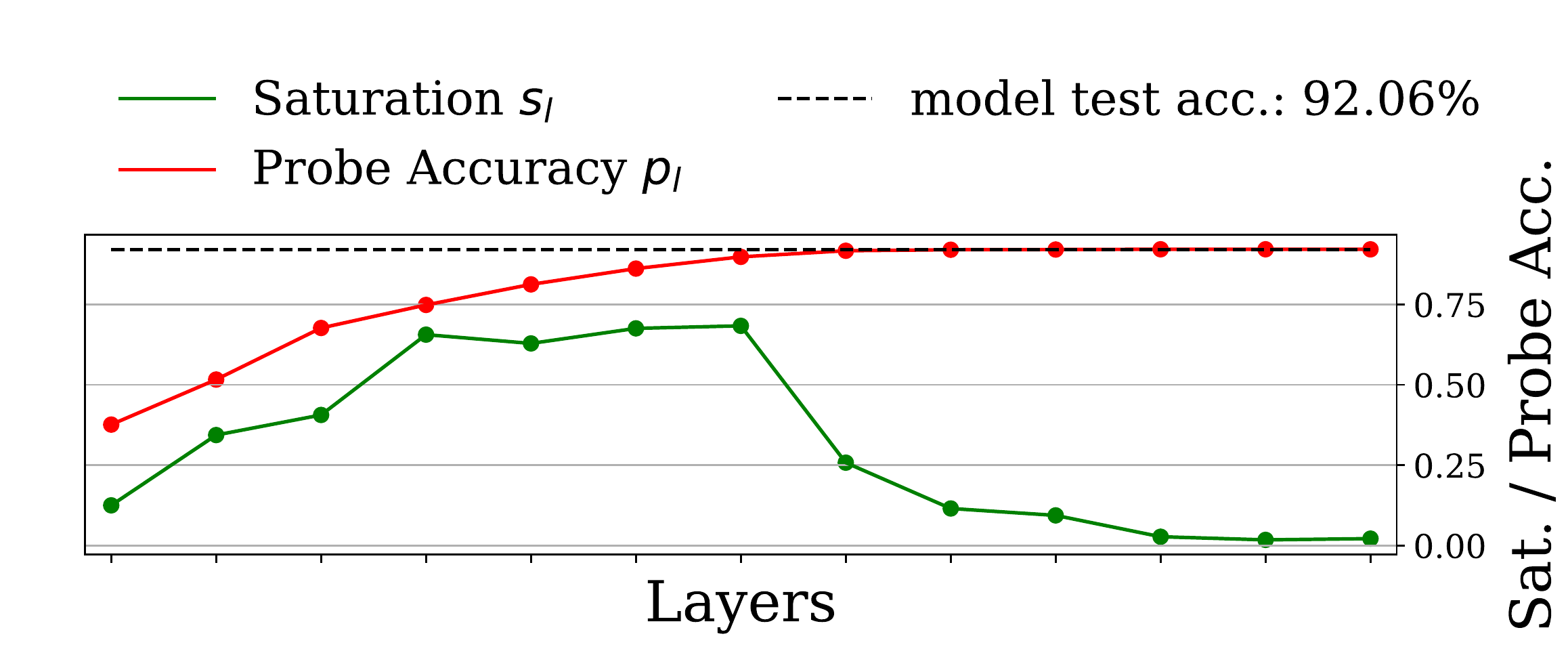}
	\caption{VGG16 exhibits a \textit{tail pattern} starting from the 8th layer. In the tail, the saturation $s_l$ is low and LRP performance $p_l$ is stagnating \cite{featurespace_saturation}, meaning that these layers are not contributing qualitatively to the prediction made by the model.}
	\label{fig:vgg16_cifar10}
\end{figure}

Saturation is another technique for expressing the activity of a convolutional layer in a single number, proposed by \ifthenelse{\value{doubleblind}>0}{Richter et al.}{us} \cite{featurespace_saturation}.
Saturation $s_l$, illustrated by the green curve in Fig.~\ref{fig:vgg16_cifar10}, is the percentage of eigendirections in the activation space of the layer $l$ required to explain $99\%$ of the variance.
Similar to the accuracy of a LRP, this results in a value bounded between $0$ and $1$.
Intuitively, saturation measures a percentage of how much the output space of a layer is ``filled'' or ``saturated'' with the data.
A sequence of low saturated layers, that is layers with $< 50\%$ of the average saturation of all other layers, is referred to as \textit{tail pattern}. 
According to \cite{featurespace_saturation}, a \textit{tail pattern} indicates that the layers belonging to the tail are not contributing qualitatively to the prediction.
This is observable in Fig.~\ref{fig:vgg16_cifar10}: The LRP accuracy of the layers in the tail stagnates; thus those layers do not advance the quality of the intermediate solution and can therefore be called unproductive layers.

%
%


Richter et al. \cite{featurespace_saturation} were able to show that such tail patterns, meaning sub-sequences of unproductive and  therewith unnecessary layers, are linked to a mismatch between input resolution and CNN architecture.

\subsection{Introducing the Receptive Field}
\label{sec:intro_rcp}
The receptive field of a convolutional layer is the area on the input image that influences the output of the unit. 
Since convolutional layers can be considered feature extractors, the receptive field size is thus the natural upper limit to the size of features that a unit in a convolutional layer can extract from the input image.
Each convolutional layer operates on the output feature map of its predecessor, thereby mapping multiple positions of its input into a single position on its output feature map.
This process results in the receptive field expanding over a sequence of convolutional layers in a CNN thereby allowing the network to detect increasingly larger features in deeper layers. 

\ifthenelse{\value{doubleblind}>0}{Richter et al.}{We} \cite{featurespace_saturation} show that this growth of the receptive field 
can lead to mismatches between the CNN architecture and the input image resolution, causing a loss of efficiency and predictive performance; both we want to avoid when designing a CNN and optimizing an existing one.
The scaling strategies for efficiently increasing the size and predictive performance of CNNs 
in EfficientNet \cite{efficientnet} and  EfficientDet \cite{efficientdet} head in a similar direction, also scaling the neural architecture in concert with the input resolution.
However, contrary to that work, we aim at a solution without an expensive grid-search.


\subsection{Computing the Receptive Field Size}
\label{sec:comp_rf}
For sequential CNNs---no multiple pathways during the forward pass---the receptive field size can be computed analytically.
We denote the receptive field size $r$ of the $l$\textsuperscript{th} layer of sequential network structure by $r_l$.
This work assumes $r_0 = 1$, which is the ``receptive field size'' of the CNN's input values.
We further assume that the input resolution $i$ and all kernels in the CNN are square shaped, allowing us to treat these values as scalars instead of $2$-tuples.
For all layers $l > 0$ in the convolutional part of a sequential network, the receptive field size can be computed with the following formula:
\begin{equation}
    r_l = r_{l-1} + ((k_{l} - 1) \prod^{l-1}_{i=0} g_{i})
\end{equation}
where $k_l$ refers to the kernel size of layer $l$ with potential dilation already accounted for and $g_{i}$ stands for the stride size of layer $i$.
The receptive field size increases with every convolutional layer $l$ with  $k_l \neq 1$.
Downsampling occurs when $g_l > 1$ and has a multiplicative effect on the growth of $r_{l+n}$.

\begin{figure}[ht]
	\centering
	\subfloat[sequence of layers used to compute $r_{l, min}$ \label{fig:arc_paths_a}] {
	    \includegraphics[width=0.8\columnwidth]{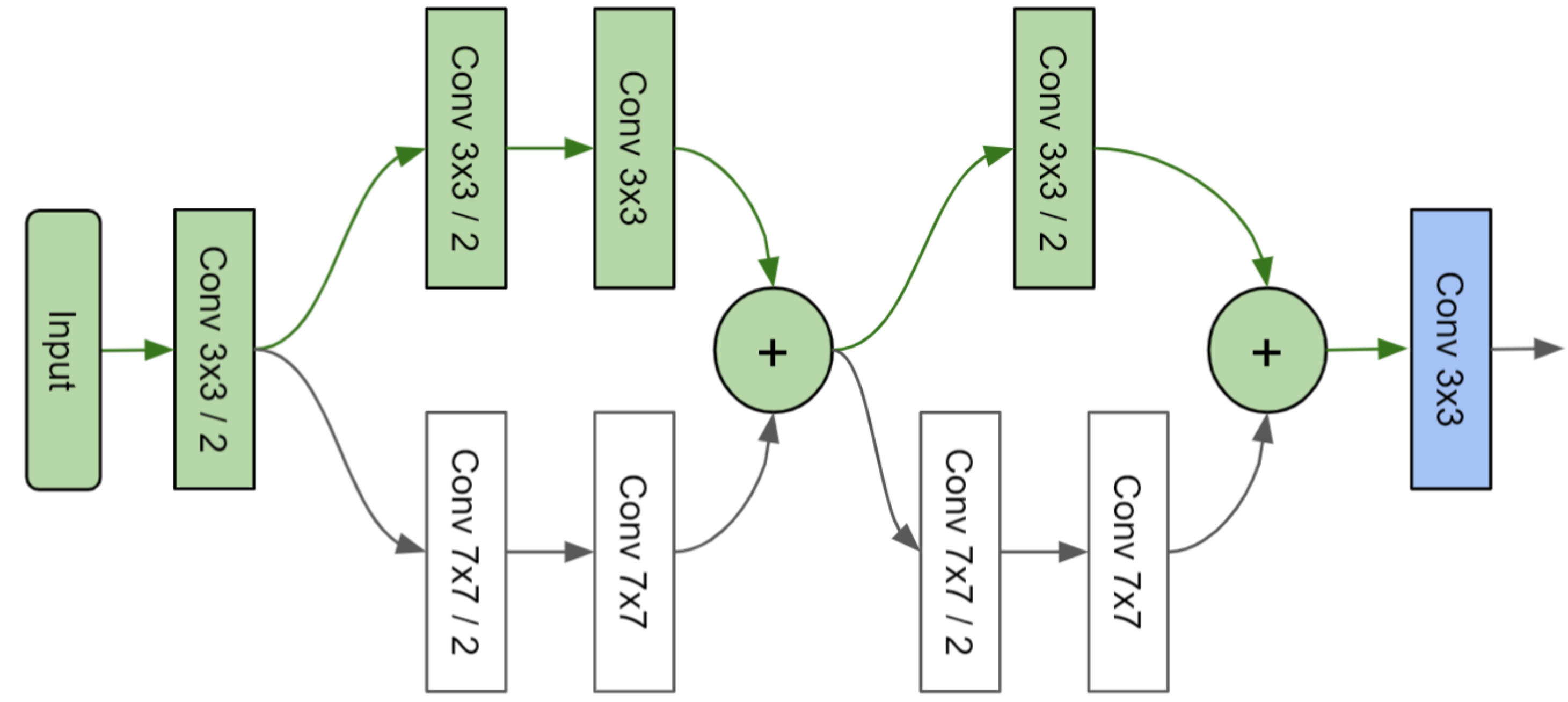} 
	}\hspace{0.3\columnwidth}
	\vspace{0.05cm}
	\subfloat[sequence of layers used to compute $r_{l, max}$ \label{fig:arc_paths_b}] {
	    \includegraphics[width=0.8\columnwidth]{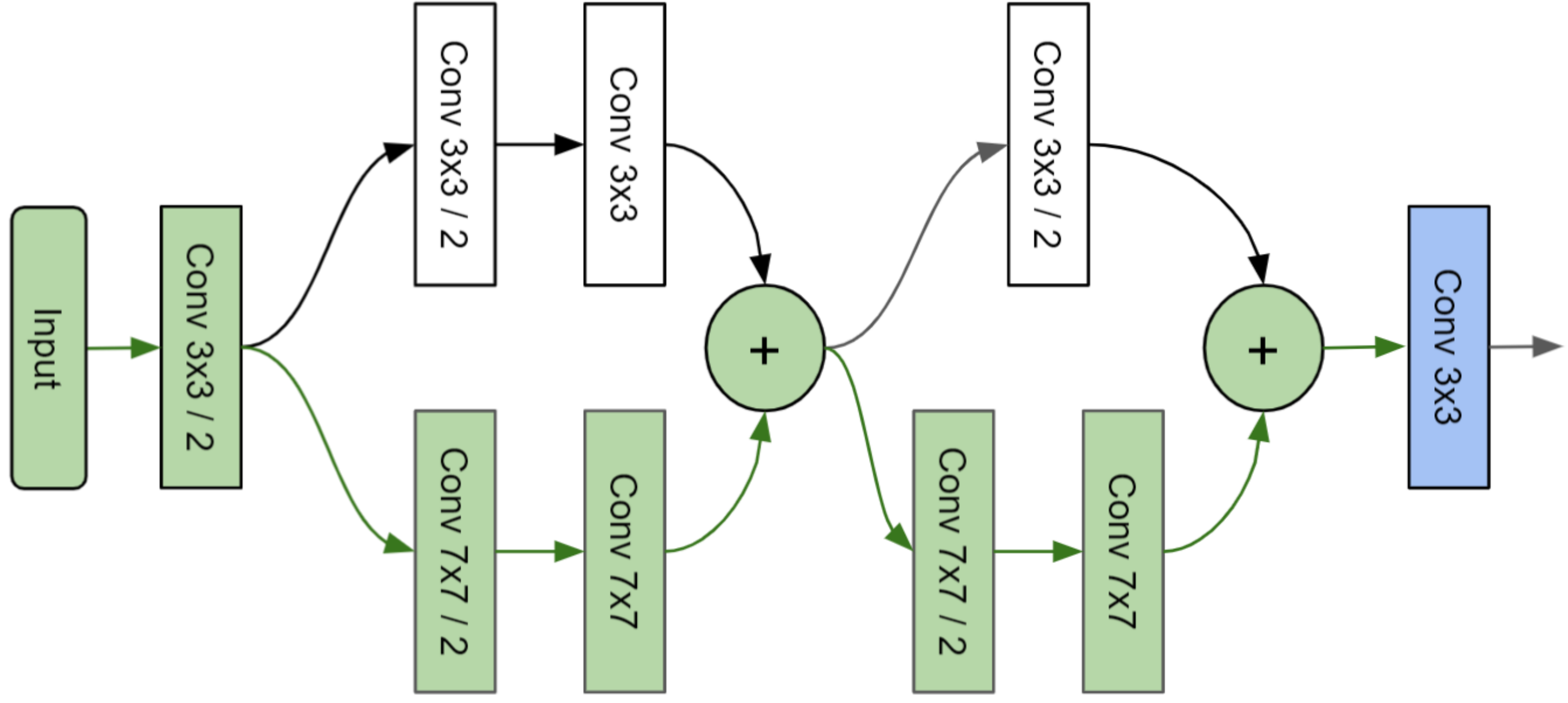}
	}
	\caption{In a non-sequential neural architecture the information in a given layer (blue) is based on multiple receptive fields with sizes in an interval $(r_{l, min}, r_{l, max})$. We can obtain the values for $r_{l, min}$ and $r_{l, max}$ analytically by computing the receptive field size of the sequences (green) with the smallest and largest receptive field size leading from the input to that layer.}
	\label{fig:arc_paths}
\end{figure}

For CNNs with non-sequential structure, that is networks with more than one computational path from the input to the output, this definition has to be refined.
Well-known examples of such  non-sequential CNNs are InceptionV3 \cite{inceptionv3} due to its parallel pathways and ResNet \cite{resnet}. 
For such networks 
all $n$ possible different receptive field sizes $r_{l_1}$ to $r_{l_n}$ for a certain layer $l$ are computed.
The maximum receptive field size of a layer $l$ refers to the largest possible field size $r_{l, \max}=\max\{r_{l_1},\ldots,r_{l_n}\}$, while the minimum size refers to
$r_{l, \min}=\min\{r_{l_1},\ldots,r_{l_n}\}$.
As shown in Fig.~\ref{fig:arc_paths}, the values for $r_{l, max}$ and $r_{l, min}$ are obtained by computing $r_l$ for the sequences of layers with the largest and smallest receptive field sizes leading to $l$.

\section{Methodology}
This section first briefly discusses the research hypothesis we will empirically investigate in section \ref{sec:experiments}.
Then we present the experimental setups used throughout this work.

\subsection{Research Hypothesis}
In section \ref{sec:tailPatterns}, we established that sequences of unproductive layers are linked to a mismatch between CNN architecture and input resolution.
In section \ref{sec:intro_rcp}, we established that the expansion of the receptive field effectively controls the size of features a layer in a CNN can extract.
Hence, the receptive fields of the CNN layers have some descriptive power over how the inference process will be distributed given a fixed input resolution.
Therefore, we hypothesize that the unproductive sequences of layers observed by Richter et al. \cite{featurespace_saturation} can be predicted using only the receptive field size $r_l$ and the input resolution $i$.
If this is the case, it should also be possible to resolve this parameter-inefficiency caused by unproductive layers by either pruning these layers or altering the receptive field expansion before training the model.

\subsection{Experimental Setups}
If not explicitly mentioned otherwise, all models are trained on the Cifar10 \cite{ cifar} dataset. 
Since our research requires many experiments and the model training  as well as LRP are a very resource-intensive process, using Cifar10 as a lightweight dataset is necessary.
Remember, Cifar10 is not a trivial task and is commonly used as a proxy problem for larger datasets like ImageNet \cite{nasnet, amoebanet}.
The CNNs used in our experiments are chosen in a way that allows to investigate specific architectural elements. 
For example, VGG16 is a model with a simple sequential structure, while ResNet18 serves to analyze the influence of skip connections.
\subsubsection{Model training} is conducted for $60$ epochs, using stochastic gradient descent with a learning rate of $0.1$ and a momentum of $0.9$ for all evaluated CNN architectures. 
The learning rate decays every $20$ epochs with a decay factor of $0.1$. The batch size chosen for this training is $64$. Preprocessing further involves channel-wise normalization using $\mu$ and $\sigma$ values taken from the original AlexNet paper \cite{alexnet}. During training, the images are furthermore randomly cropped and horizontally flipped with a $50\%$ probability.

\subsubsection{Saturation} is computed during training of the final epoch on every convolutional and fully connected layer using a $\delta$ of $99\%$\ifthenelse{\value{doubleblind}>0}{.}{, which is the standard configuration recommended by us \cite{featurespace_saturation}.} 

\subsubsection{LRP} accuracy values are computed on the same layers as saturation after the model training.
By training the LRPs, this work differs from the procedures of Alain and Bengio \cite{probes} by not global-pooling the feature maps to a single vector to avoid artifacts caused by the aggressive downsampling.
Based on an ablative study on LRP performance \cite{featurespace_saturation}, large feature maps to a size of $4 \times 4$ pixels are adaptive average pooled, which is a good compromise between computational feasibility and reliability of the accuracy obtained from the LRP.

\section{Experiments}
\label{sec:experiments}

We investigate our research hypothesis, that unproductive layers can be predicted using the receptive field size and input resolution, in multiple steps.
First, we test it on simple, sequential neural architectures. 
Based on these findings, we expand our hypothesis on multi-path architectures and on architectures with residual connections.
We then show that attention-mechanisms in the model do not influence the previously made observations.

\begin{figure*}[htb!]
	\centering
	\subfloat[VGG11---border layer at conv6 \label{fig:vgg_receptive_field_VGG11}]{
	    \includegraphics[width=0.9\columnwidth,trim=0mm 2mm 0mm 12mm, clip]{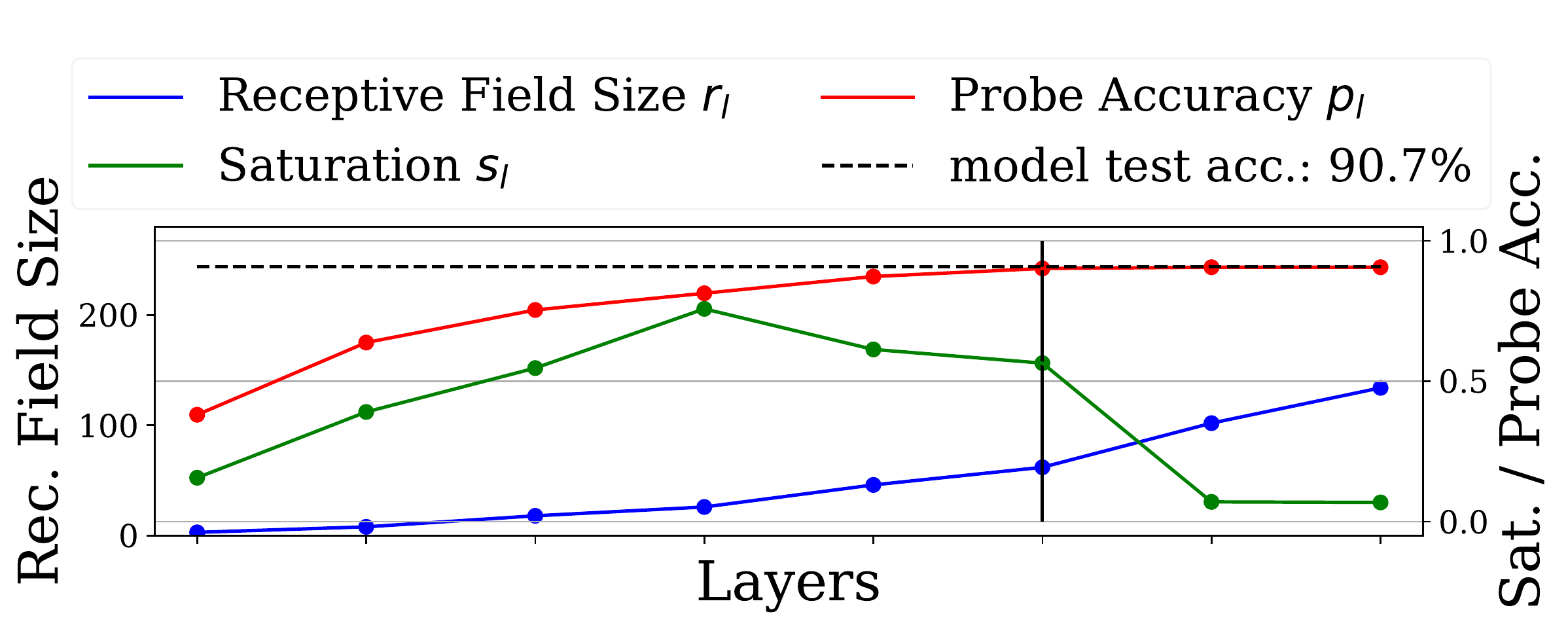}
	}
	\hspace{0.1\columnwidth}
	\subfloat[VGG16---border layer at conv8\label{fig:vgg_receptive_field_VGG16}]{
	    \includegraphics[width=0.9\columnwidth,trim=0mm 2mm 0mm 12mm, clip]{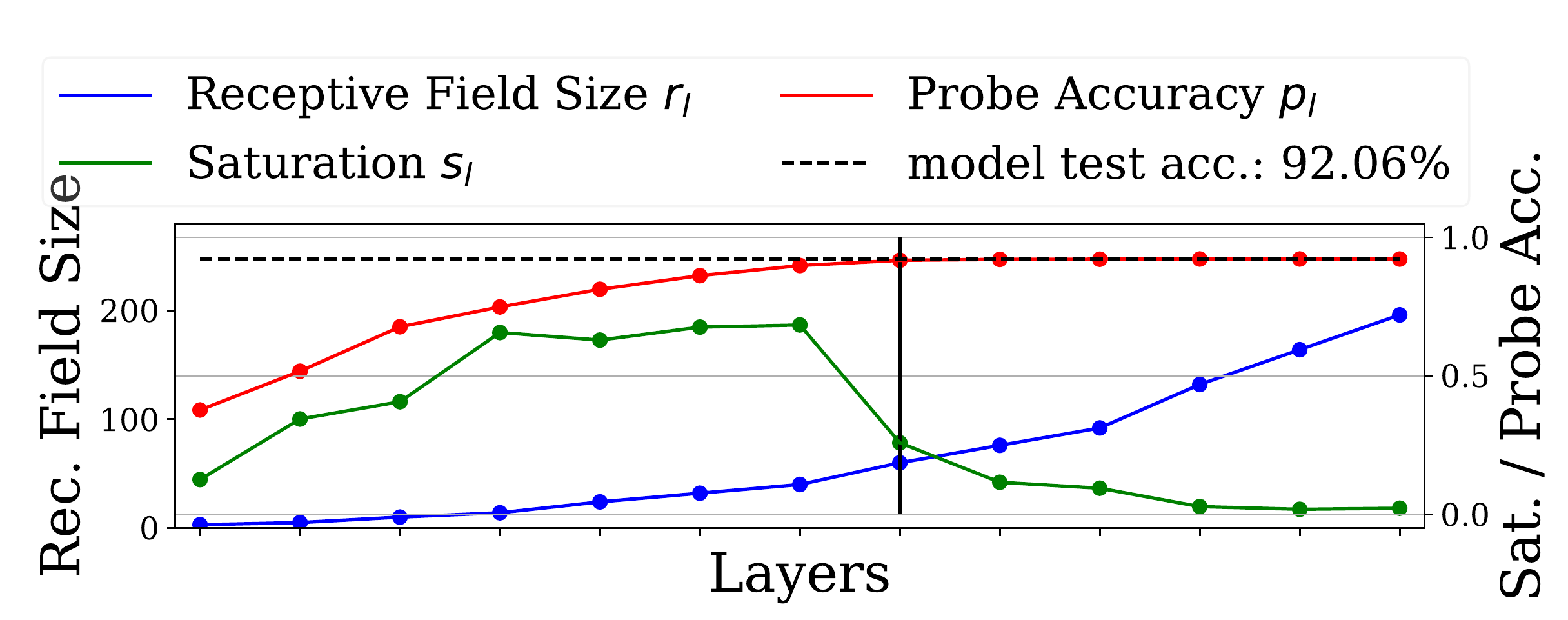}
	}
	\vspace{.1cm}\qquad
	\subfloat[VGG13---border layer at conv8\label{fig:vgg_receptive_field_VGG13}]{
	    \includegraphics[width=0.9\columnwidth,trim=0mm 2mm 0mm 12mm, clip]{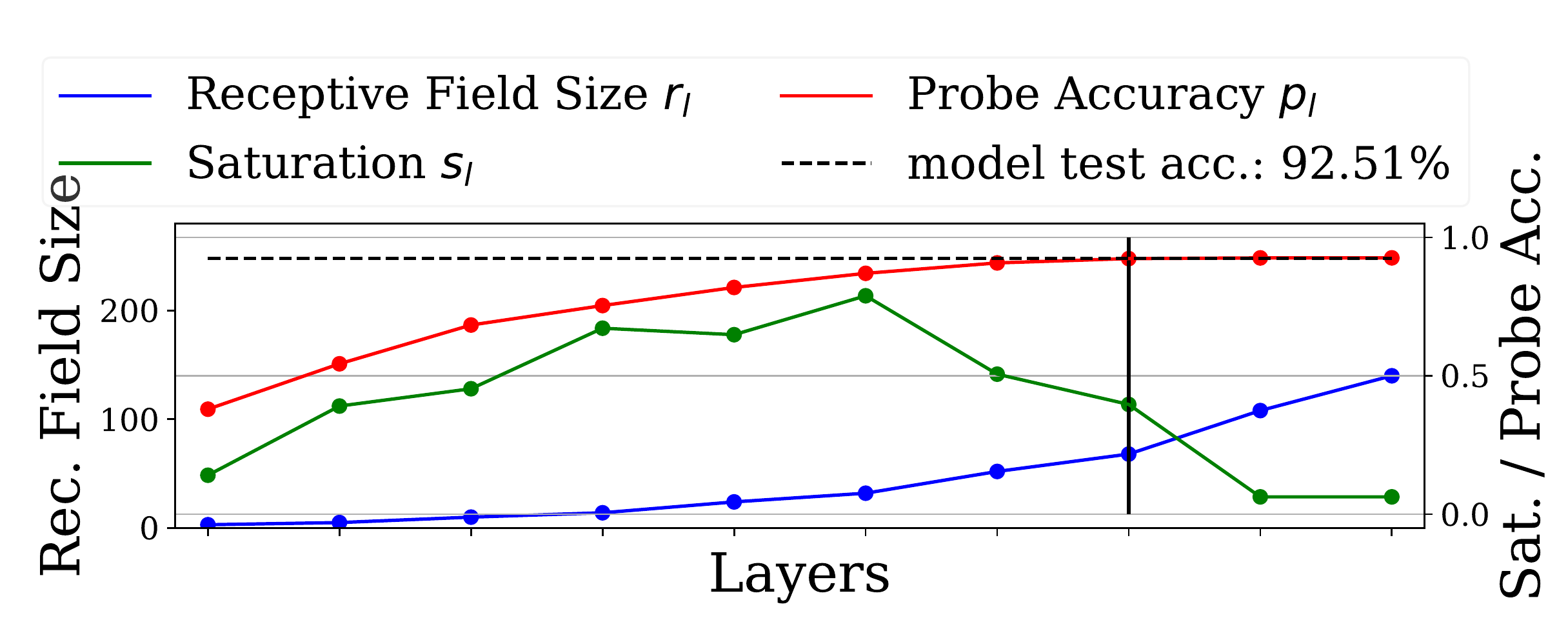}
	}
	\hspace{0.1\columnwidth}
	\subfloat[VGG19---border layer at conv8\label{fig:vgg_receptive_field_VGG19}] {
	    \includegraphics[width=0.9\columnwidth,trim=0mm 2mm 0mm 12mm, clip]{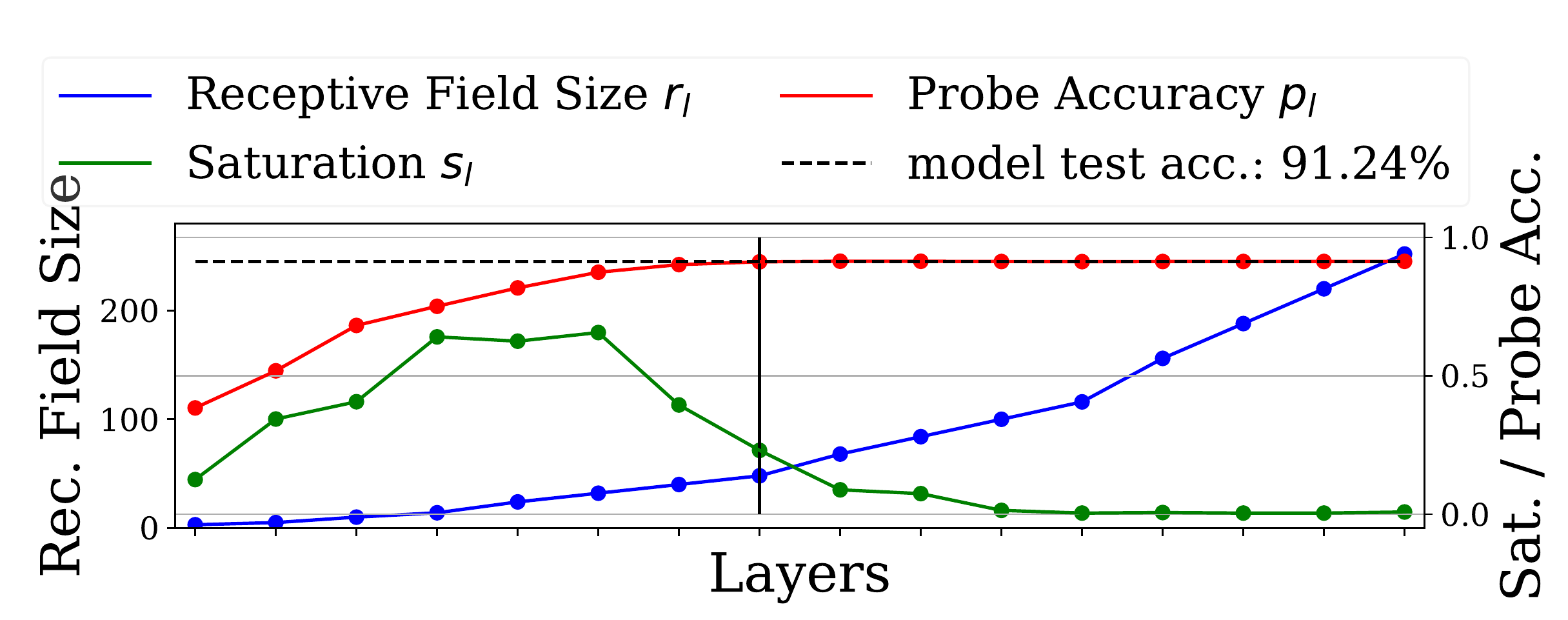}
	}
	\caption{By analyzing the receptive field size, the start of the unproductive convolutional layers within the network architecture can be predicted on VGG-architectures of varying depth. The border layer, marked with a black bar, separates the part of the network that contributes to the quality of the prediction from the part that does not. The  border layer is the first layer with a $r_{l-1} > i$, where $i$ is the maximum value of either the height and width of the input image. Here, Cifar10 with the resolution of $32\times32$ pixel is used, therefore $i=32$.}
	\label{fig:vgg_receptive_field}
\end{figure*}

\subsection{border layer for Sequential CNN Architectures}
\label{sec:seq}
\begin{figure*}[htb!]
	\centering
	\subfloat[VGG11---border layer at conv6]{
	    \includegraphics[width=0.9\columnwidth,trim=0mm 2mm 0mm 12mm, clip]{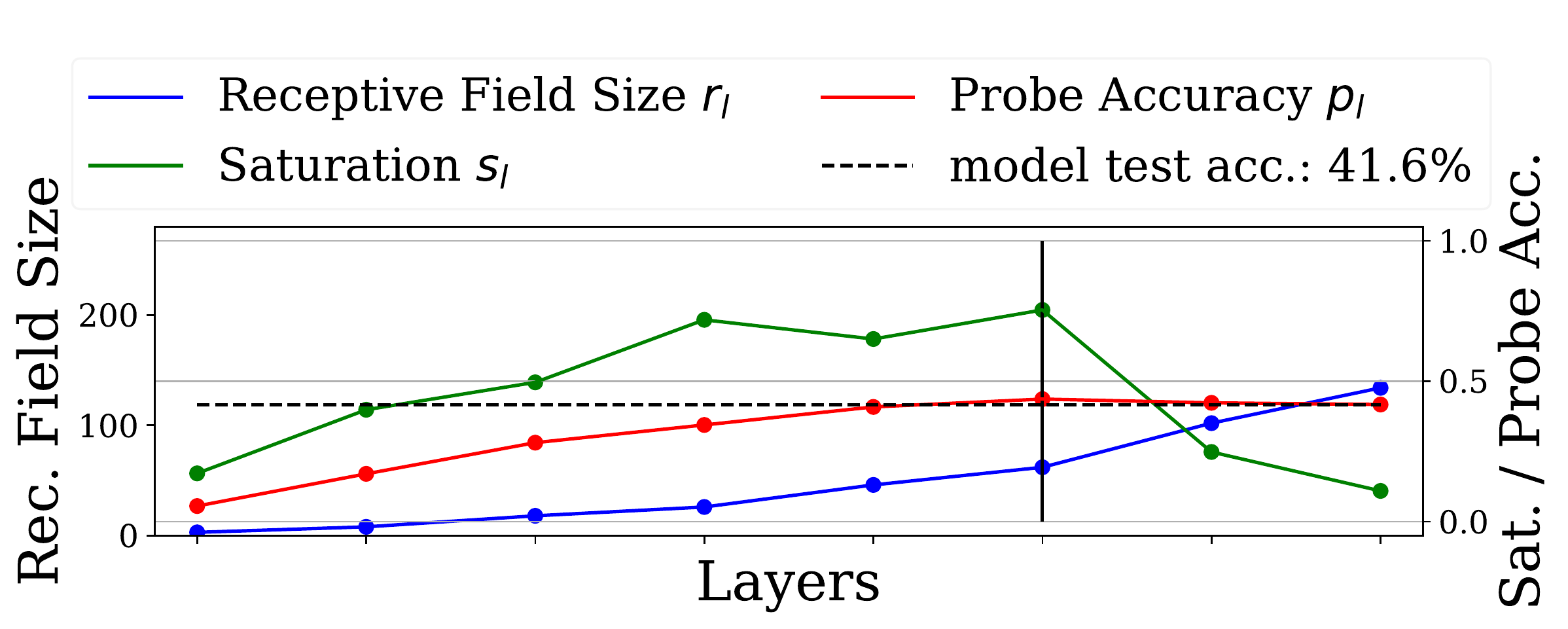}
	}\hspace{0.1\columnwidth}
	\subfloat[VGG16---border layer at conv8]{
	    \includegraphics[width=0.9\columnwidth,trim=0mm 2mm 0mm 12mm, clip]{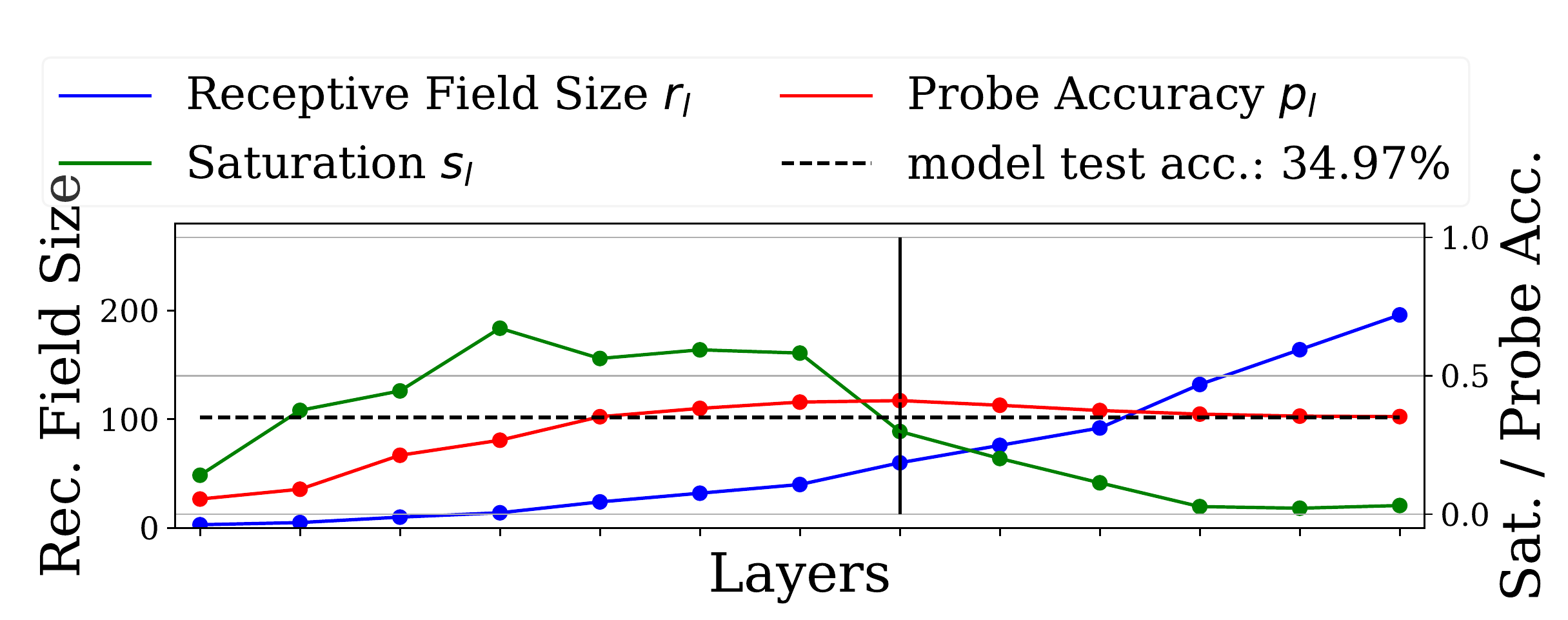}
	}
	\caption{The prediction by the border layer shown in Fig.~\ref{fig:vgg_receptive_field} holds when changing the training data set to TinyImageNet \cite{tinyimnet}, as long as the input resolution stays at $32\times32$ pixels.} 
	\label{fig:vgg16_tinyimgnet}
\end{figure*}

This section starts the investigation by analyzing to which extent it is possible to predict unproductive layers in a simple sequential convolutional neural architecture, that is a sequence of convolutional and pooling layers.
As established in section \ref{sec:comp_rf} the receptive field size can be computed unambiguously in this case.
The VGG-family of CNN architectures by Simonyan and Zisserman \cite{vgg} is exemplary for sequential CNN architectures and is consequently used in the experiments,
cf.\ Fig.~\ref{fig:vgg_receptive_field} and Fig. \ref{fig:vgg16_tinyimgnet}.

For answering our research question, whether it is possible to predict unproductive subsequences of layers, we hypothesize that layers become unproductive if they cannot integrate novel information into a single feature map position.
This is the case when $r_{l-1} > i$, where $r_{l-1}$ is the receptive field of the layer $l$`s input and $i$ is the input resolution.
Since the receptive field size grows monotonically in a simple sequential architecture, a clear border separating the productive part of the model from the unproductive part based on the condition $r_{l-1} > i$ can be defined.
The first layer $l$ with the condition $r_{l-1} > i$ is referred to as border layer $b$, since it effectively separates the productive from the unproductive layers of the CNN.
By training VGG11, VGG13, VGG16, and VGG19 on the Cifar10 dataset using its native image resolution  of $32\times 32$ pixels, the receptive field size, the saturation values, and the LRP accuracy are computed.

As depicted in Fig.~\ref{fig:vgg_receptive_field}, the saturation and the LRP accuracy of all four VGG setups increases significantly up to the border layer, after the border layer the LRP accuracy no longer improves.
Consequently, the border layer separates the convolutional layers that contribute qualitatively, i.e., by the increasing LRP accuracy and a high saturation value, from convolutional layers that do not.
The observed behavior is also reproducible when training the models on TinyImageNet, while using the same setups and image resolution as plotted in Fig.~\ref{fig:vgg16_tinyimgnet}.
The plots of Fig.~\ref{fig:vgg_receptive_field} and  \ref{fig:vgg16_tinyimgnet} show that the border layer behavior is consistent over different network depths of the VGG-family and does not depend on the dataset used.

We conduct two experiments with different sequential architectures trained on Cifar~10 to investigate whether the observed border layer behavior can be reproduced. 
The first experiment uses a modified ResNet18 \cite{resnet} architecture.
In this modified version, all skip connections of ResNet18 are disabled, ensuring a sequential network architecture. 
This sequential ResNet18 architecture model differs from VGG-style models in numerous ways.
Firstly, BatchNorm \cite{batchnorm} and strided convolutions are used for downsampling instead of MaxPooling layers. Furthermore,  the sequential ResNet18 features a stem consisting of two consecutive downsampling layers at the input of the model, which strongly affect the growth of the receptive field size.
The second experiment utilizes a modified VGG19 with dilated convolutions, increasing the kernel sizes to $7 \times 7$.

\begin{figure*}[tb!]
	\centering
	\subfloat[ResNet18; no skip connections---border layer at conv5 \label{fig:vgg16resnetnoskipRestNet18}]{
	    \includegraphics[width=0.9\columnwidth,trim=0mm 2mm 0mm 12mm, clip]{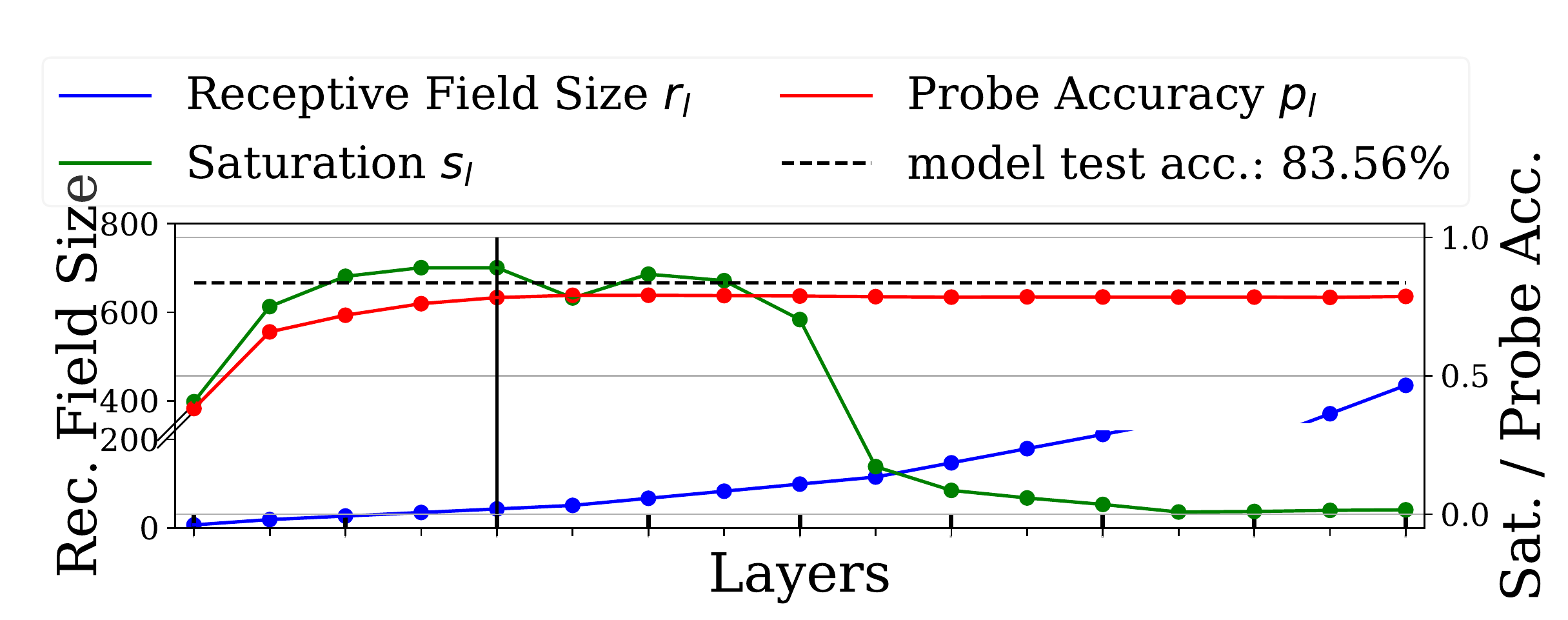}
	} \hspace{0.1\columnwidth}
	\subfloat[VGG19; dilation=3---border layer at conv5 \label{fig:vgg16resnetnoskipModVGG19}]{
	    \includegraphics[width=0.9\columnwidth,trim=0mm 2mm 0mm 12mm, clip]{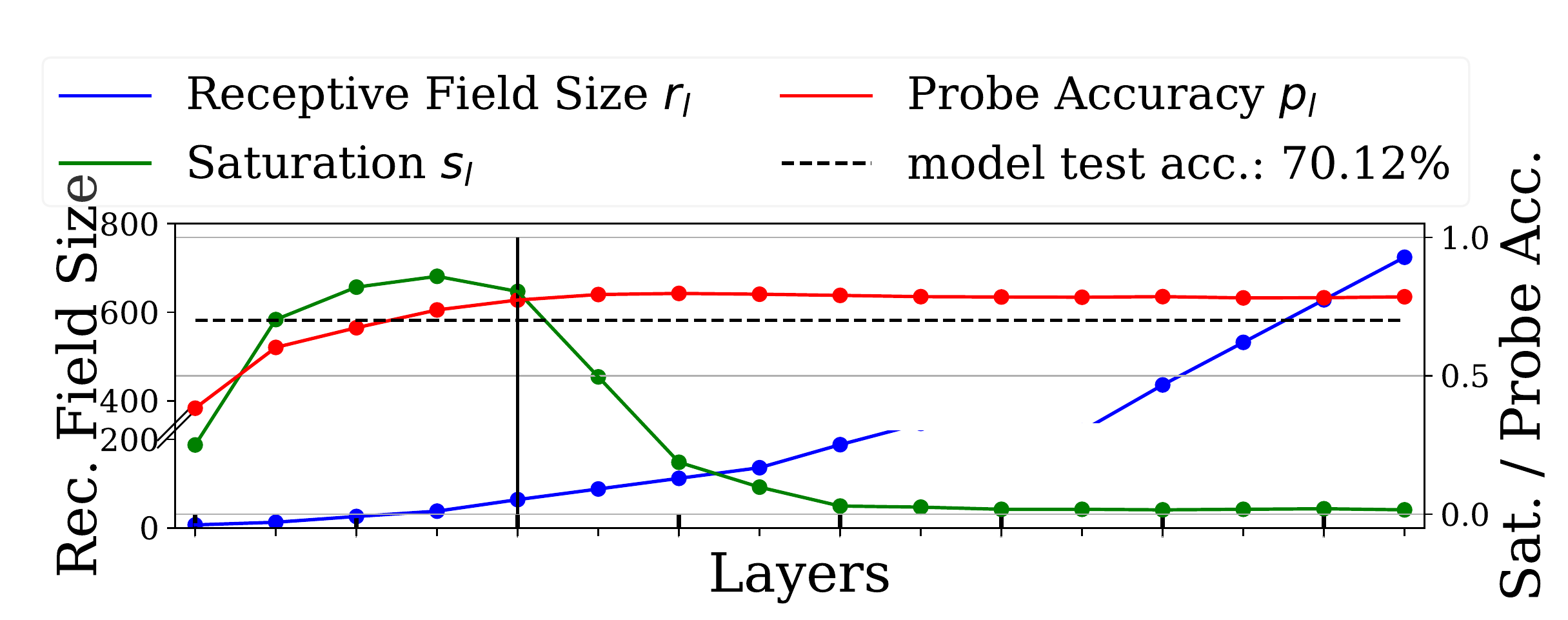}
	}
	\caption{Predicting unproductive layers with the border layer is also feasible for other architectures.
	ResNet18 with disabled skip connections \protect\subref{fig:vgg16resnetnoskipRestNet18} uses two initial downsampling layers. The depicted VGG19 variant \protect\subref{fig:vgg16resnetnoskipModVGG19} uses dilated convolutions. Both properties affect the growth of the receptive field size differently compared to the models in Fig.~\ref{fig:vgg_receptive_field} and \ref{fig:vgg16_tinyimgnet}.} 
	\label{fig:vgg16resnetnoskip}
\end{figure*}

In line with the border layer behaviors,  both experiments shown in 
Fig.~\ref{fig:vgg16resnetnoskip}  prove that the border layer separates the productive layers from the unproductive ones. The observation is thus consistent with previous experiments of Fig.~\ref{fig:vgg_receptive_field} and  \ref{fig:vgg16_tinyimgnet}.

%

\subsection{border layer for Non-Sequential CNN Architectures}
\label{sec:multipath}

\begin{figure}[hb!]
	\centering
	\subfloat[Overall structure\label{fig:mp_arc_overview}]{
	    \includegraphics[width=0.35\columnwidth]{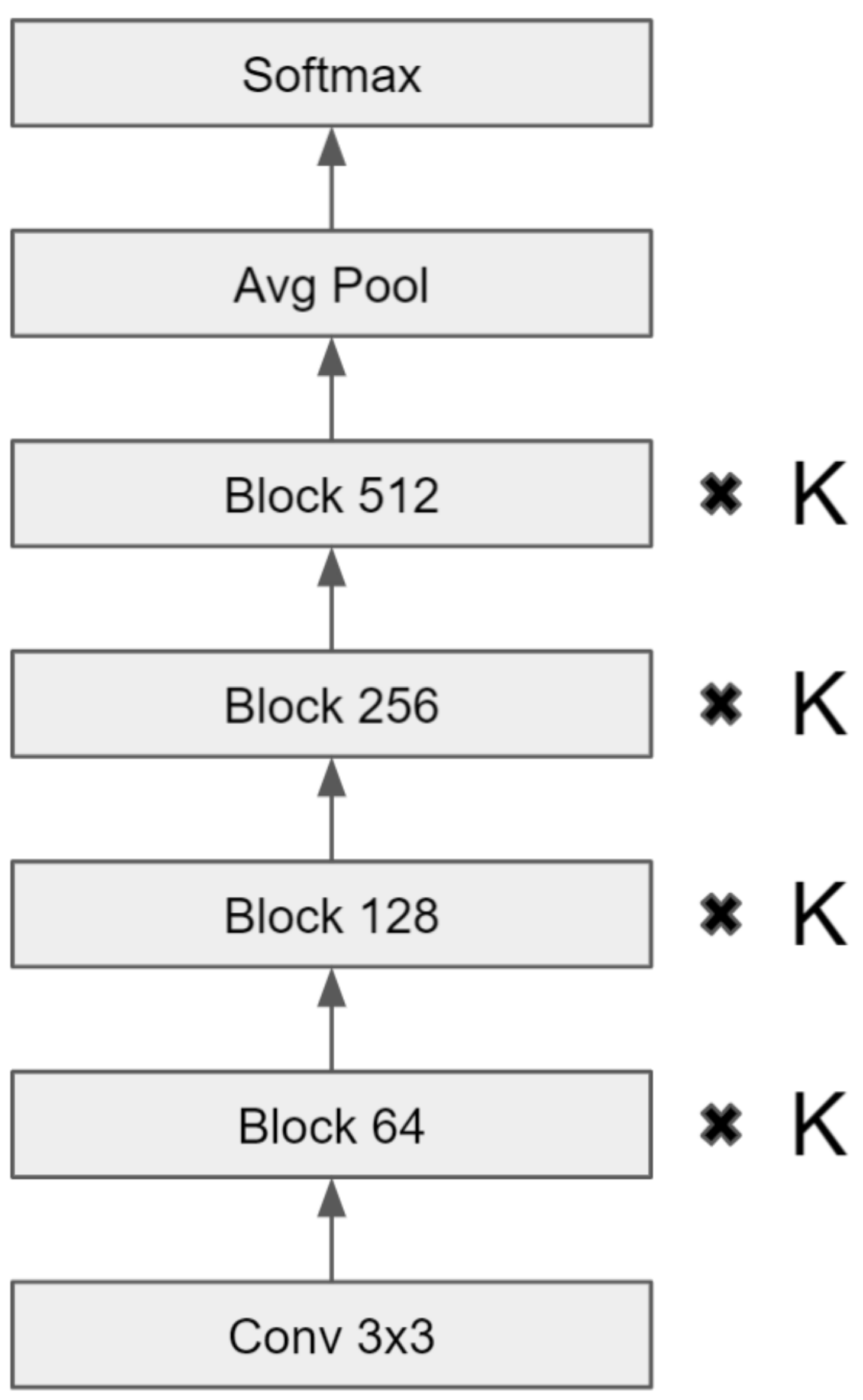}
	}\quad
	\subfloat[Module A\label{fig:mp_arc_modA}]{
	    \includegraphics[width=0.35\columnwidth]{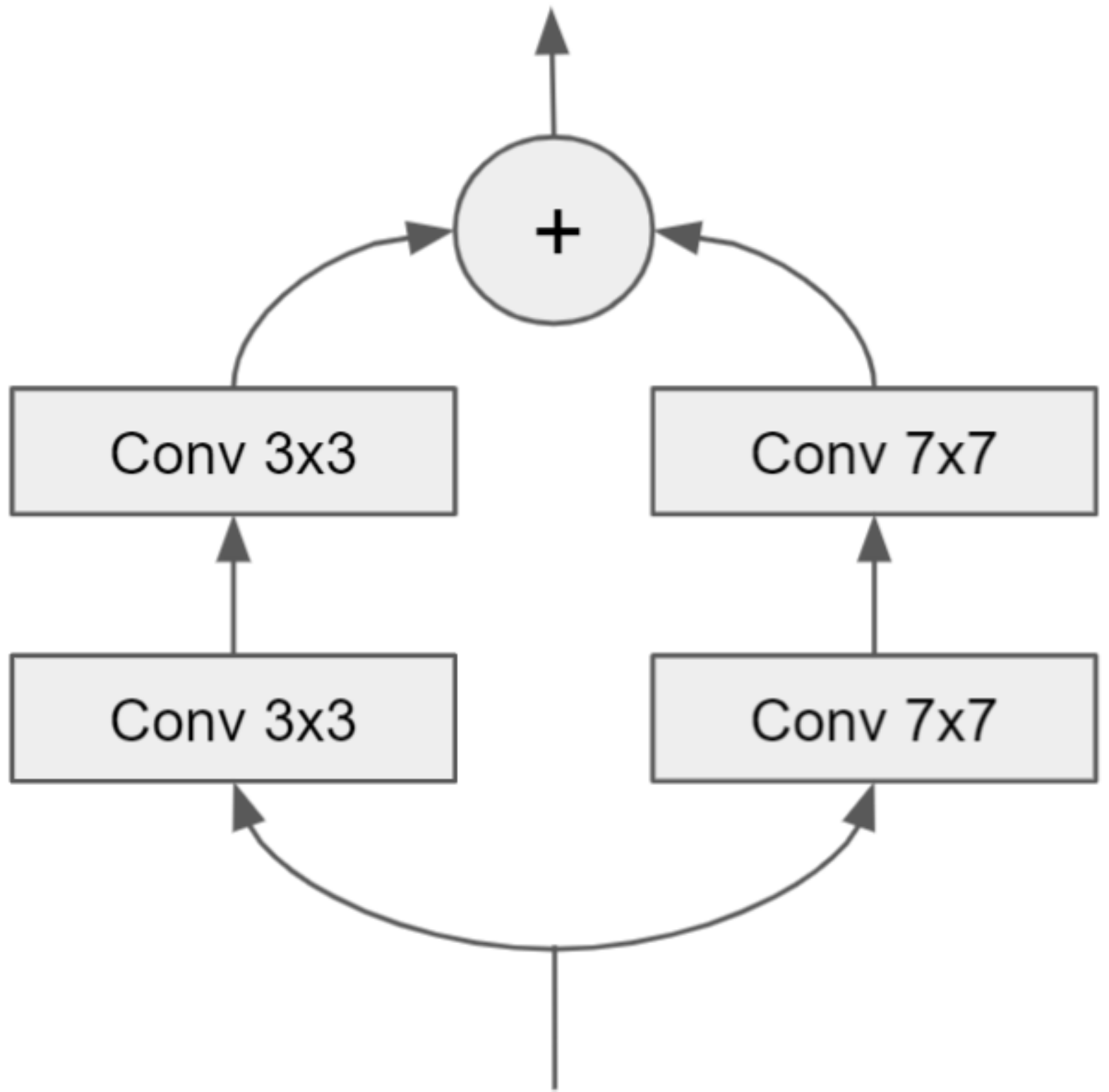}
	}
		\subfloat[Module B\label{fig:mp_arc_modB}]{
	    \includegraphics[width=0.35\columnwidth]{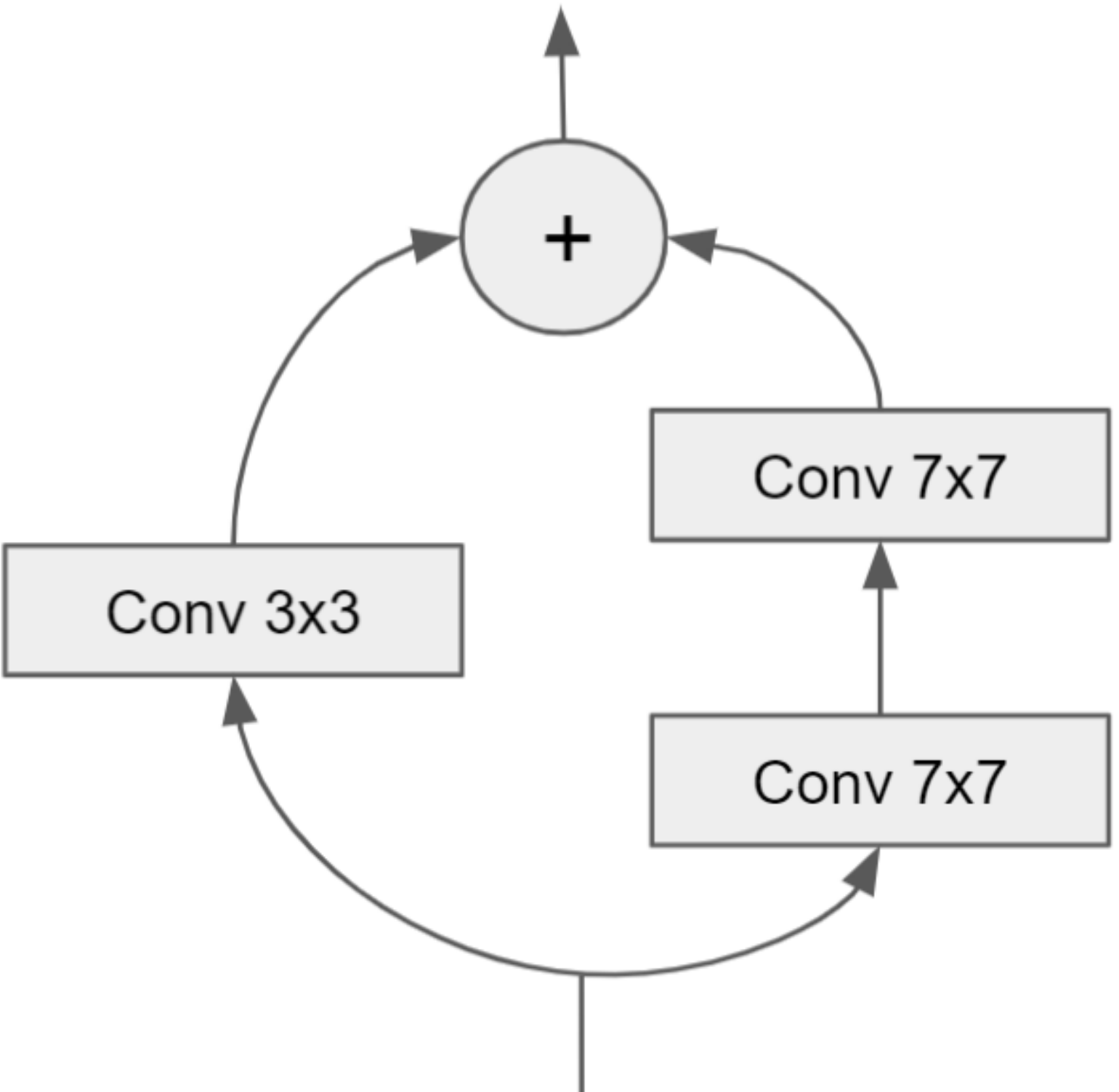}
	}
	\caption{The architecture used for the experiments in this section. Each stage consists of $k$ blocks and has double the filters of the previous stage. The first layer of each stage is a downsampling layer with a stride size of 2.
	The building block for the shallow architecture MPNet18 is depicted in \protect\subref{fig:mp_arc_modA}, the building block of the deep architecture MPNet36 is depicted in \protect\subref{fig:mp_arc_modB}.}
	\label{fig:mp_arc}
\end{figure}

This subsection investigates if predicting unproductive sequences of convolutional layers is still possible even when the network structure is non-sequential. 
In comparison to sequential CNNs, in multipath architectures information based on multiple receptive field sizes may be present in any layer's input.
To observe whether changes to the number of layers and a more substantial deviation in the receptive field sizes within a module affect the distribution of the inference process in unexpected ways, we design a multipath ``model organism`` architecture that is also simple to analyze.
%
The designed generic non-sequential multipath architecture is depicted in Fig.~\ref{fig:mp_arc} \subref{fig:mp_arc_overview}.

Its design follows the conventions regarding downsampling and general structure utilized in various architectures such as ResNet \cite{resnet}, AmoebaNet \cite{amoebanet} and EfficientNet \cite{efficientnet}. This non-sequential architecture has four stages consisting of building blocks with similar filter sizes. The first layer in each stage is a downsampling layer that reduces the size of the feature maps by having a stride size of 2. For generating variances, two distinct architectures---a shallow and a deep architecture---are defined.
The shallow multipath architecture, called  MPNet18, uses two Module A building blocks per stage. The Module A building blocks, illustrated in Fig. ~\ref{fig:mp_arc} \subref{fig:mp_arc_modA}, consist of a $3 \times 3$ convolutional path and a $7 \times 7$ convolutional path, merged by an element-wise addition. 
The second, deep multipath architecture, called  MPNet36, uses four Module B building blocks. As shown in Fig.~\ref{fig:mp_arc} \subref{fig:mp_arc_modB}, Module B building blocks have a different number of layers in each pathway and thus a larger difference between $r_{l, min}$ and $r_{l, max}$.

\begin{figure*}[tb!]
	\centering
	\subfloat[MPNet18---border layer $b_{min}$ at conv11 and $b_{max}$ at conv7]{
	    \includegraphics[width=0.95\columnwidth,trim=0mm 2mm 0mm 8mm, clip]{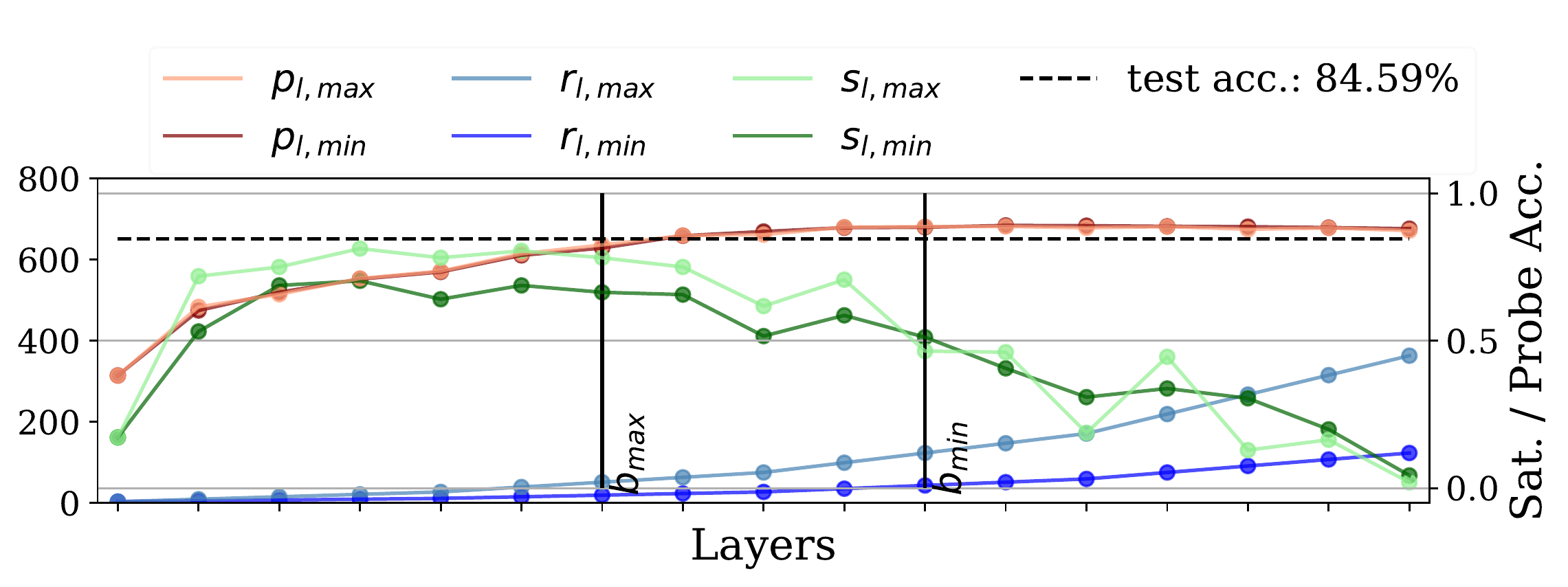}
	} \hspace{0.05\columnwidth}
	\subfloat[MPNet36---border layer $b_{min}$ at conv22 and $b_{max}$ at conv7]{
	    \includegraphics[width=0.95\columnwidth,trim=0mm 2mm 0mm 8mm, clip]{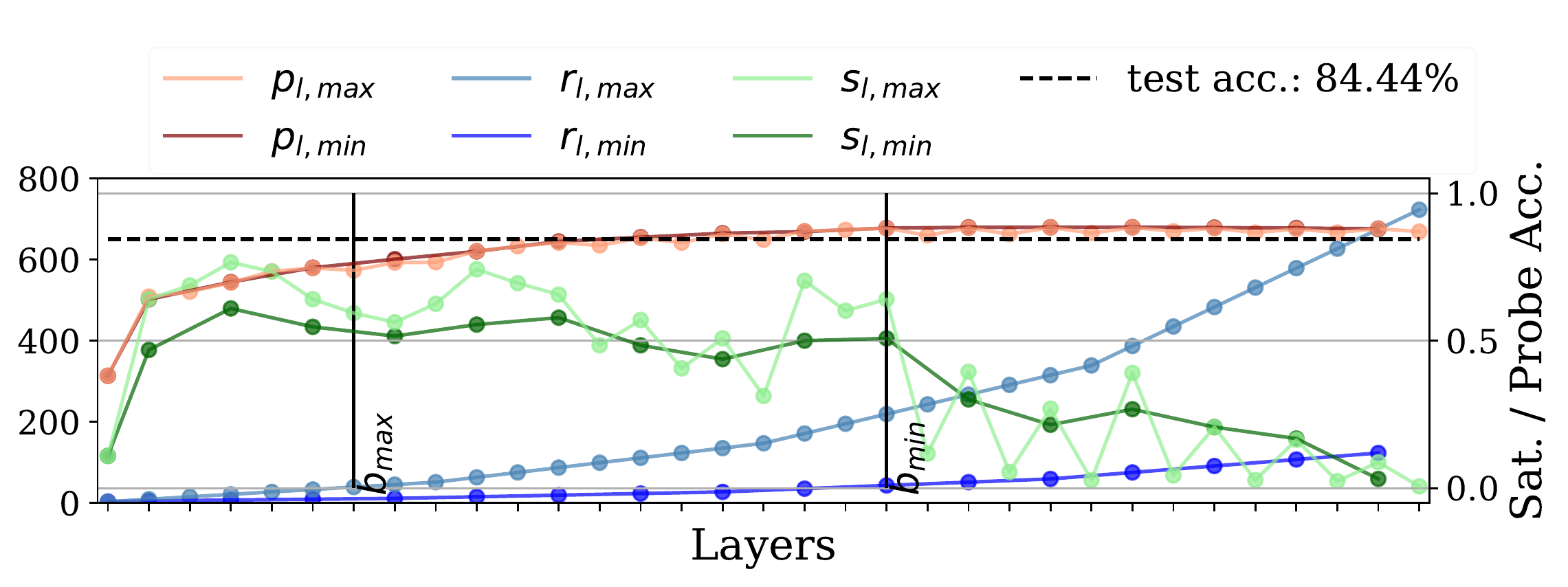}
	}
	\caption{The border layers $b_{min}$ divides the productive and unproductive layers for non-sequential networks ---similar to the border layer in sequential architectures. MPNet18 is our proposed shallow multipath network, MPNet36 is our proposed deep multipath network.}
	\label{fig:mp_analysis}
\end{figure*}

In MPNet18 and MPNet36, the convolutional layers use same-padding, batch normalization, and ReLU-activation functions.
For merging the pathway, element-wise addition is used since it does not increase the number of filters as concatenation does.
Merging by element-wise addition avoids $1 \times 1$ convolutions for dimension reduction, which could induce noisy artifacts into the analysis. Having only two distinct pathways with distinct kernel sizes make it easy to compute $r_{l, min}$ and $r_{l, max}$.
The two-pathway design further allows us to view the model as two sequences with drastically different expansion of $r_{l, max}$, which greatly simplifies the visualization of the architectures in Fig.~\ref{fig:mp_analysis}.
%
Like the sequential CNN, MPNet18 and MPNet36 are trained and evaluated within the same experimental setup on Cifar10. 

Fig.~\ref{fig:mp_analysis} shows that in the shallow as well as in the deep multipath architecture, the LRP accuracy performs very similarly.
Remarkable is that the border layer $b_{max}$ based on the largest receptive field size has no apparent effect on the development of the LRP accuracy.
However, the border layer $b_{min}$ of the smallest receptive field size exhibits the same behavior that was observed for the border layers of Fig.~\ref{fig:vgg_receptive_field} to \ref{fig:vgg16resnetnoskip}.
The behavior of the  \textit{minimum border layer} $b_{min}$ supports the notion that the integration of novel information is critical for the improvement of the solution.
However, it also shows that the network will not greedily integrate all available information as soon as possible into a single position on the feature map.

\subsection{Skip Connections Allow Qualitative Improvements Past the border layer}
\label{sec:residual}

\begin{figure*}[t!]
	\centering
	\subfloat[ResNet18---border layer at conv11]{
	    \includegraphics[width=0.95\columnwidth,trim=0mm 2mm 0mm 12mm, clip]{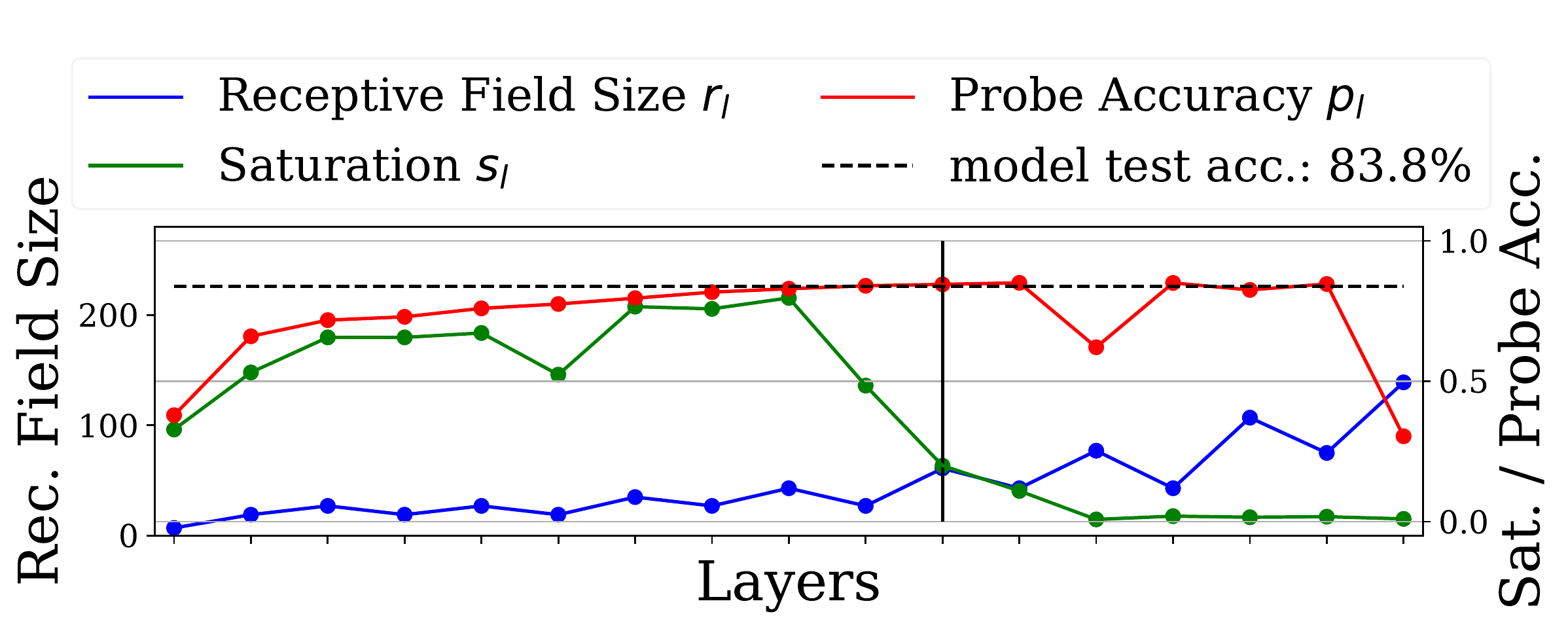}
	} \hspace{0.05\columnwidth}
	\subfloat[ResNet34---border layer at conv17]{
	    \includegraphics[width=0.95\columnwidth,trim=0mm 2mm 0mm 12mm, clip]{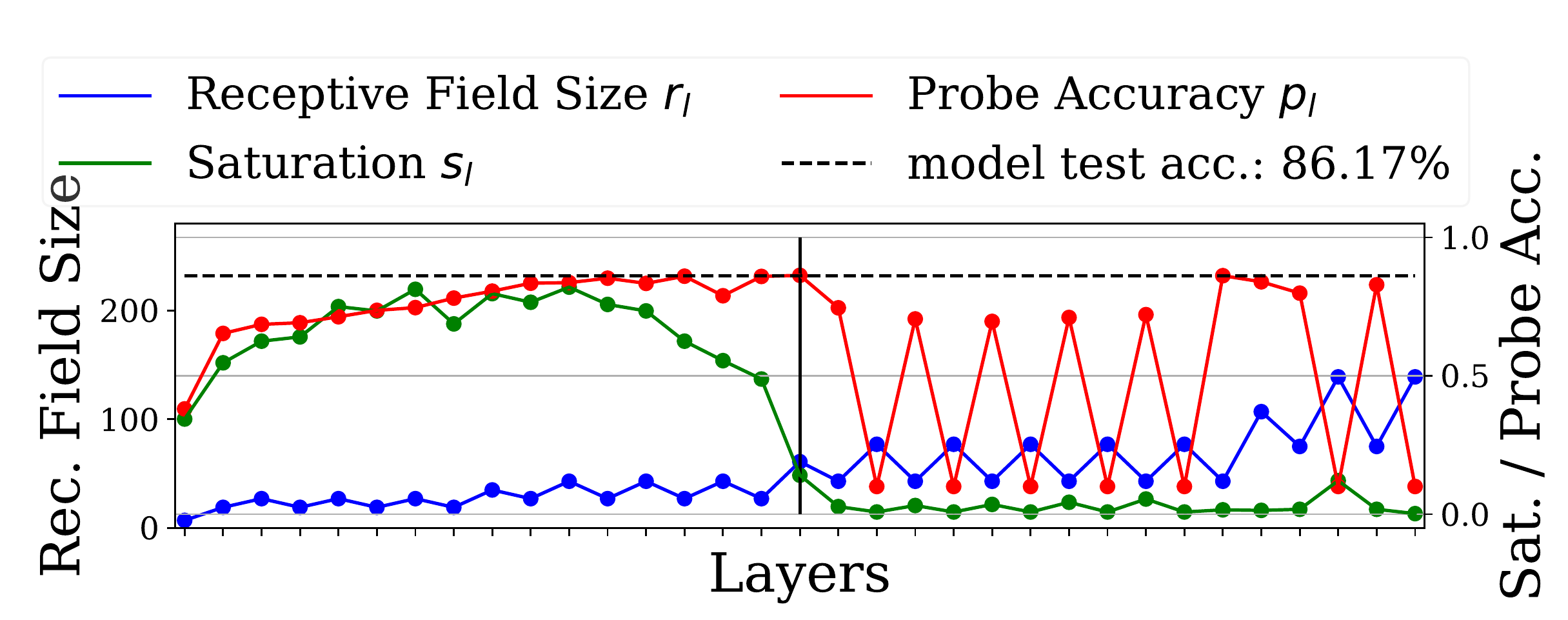}
	}
	\caption{The border layer computed from the smallest receptive field size $b_{min}$ can predict unproductive layers for ResNet architectures. 
	The skip connections allow information based on smaller receptive field sizes to skip layers, resulting in a later border layer $b_{min}$ compared to the same network with disabled skip connections. This allows networks with skip connections to involve more layers in the inference process than it would be possible compared to a simple sequential architecture with a similar layout. For an example compare Fig.~\ref{fig:vgg16resnetnoskip} (a) with Fig.~\ref{fig:receptive_field_resnet} (a). }
	\label{fig:receptive_field_resnet}
\end{figure*}

Skip connections are a special case of non-sequential architectures since they effectively create pathways that do not expand the size of the receptive field and are often parameterless. 
Multiple variants of skip connections \cite{resnet, highway, densenet} have been proposed over the years, which generally deviate in the way the pathways are merged and whether the skip connections themselves are parameterizable.
From the variants of skip connections, the skip connection proposed by the authors of He et al. \cite{resnet} is the most common type used. Next to ResNet, this variant can be found in architectures like AmoebaNet \cite{amoebanet}, MobileNet \cite{mobilenetv2, mobilenetv3}, and EfficientNet \cite{efficientnet}. Due to this variant's frequent use, this work focuses on this particular skip connection, which uses an element-wise addition for merging the pathways and contains no trainable parameters. 

If the receptive field expansion behavior is consistent with the behavior of multi-path architectures discussed in the Section \ref{sec:multipath}, the border layer $b_{min}$  should separate the unproductive tail of convolutional layers.
We examine this idea on the ResNet18 and ResNet34 architectures.

The plots in Fig.~\ref{fig:receptive_field_resnet} indicate that the border layer assumption also applies to non-sequential architectures with skip connections.
On both tested ResNet architectures, the qualitative improvement of the predictive performance stops when the border layer is reached.
The zig-zag-pattern of the minimal receptive field size are caused by the fact that skip connections effectively allow the CNN to skip all layers except the stem. Thus, the receptive field size is reduced effectively at each merging block of the pathways. 
In later layers of the ANN, amplitudes of this zig-zag pattern become more pronounced, caused by the additional downsampling layers that increase the growth rate of the receptive field size.
The LRP accuracy of ResNet18 and ResNet34 also shows some anomalies, in the form of sudden drops of its value, occurring exclusively after the border layer.
These drops of the LRP accuracy result from skipped layers because the CNN does not utilize layers that are bypassed by skip connections.
This behavior at the bypassed layers is identical with the observations by  Alain and Bengio \cite{probes} using linear classifier probes. 

\subsection{Attention Mechanisms Do Not Influence LRP and Saturation Values}\label{sec:attention}
Building on the findings from Section \ref{sec:seq} to \ref{sec:residual}, the working hypothesis is that there are no influences on the LRP and saturation values by any attention mechanisms.
Thus, attention mechanisms are considered as simple layer add-ons generating dynamical weights based on the input.
Squeeze-and-Excitation (SE) modules  \cite{Hu2018} are a filter-wise attention mechanism, since the weighting is applied on each feature map.
In contrast to SE modules, a spatial attention mechanism applies a single weight on each feature map position.  The Convolutional Block Attention Module (CBAM) \cite{cbam} is a combination of both.
In all cases, one or multiple auto-encoder-like structures are condensing the entire stack of feature maps into a set of weights, usually, with some global pooling strategy \cite{networkinnetwork}.
The attention mechanism thereby incorporates global information about the image into the feature map via multiplication and hence changes the information present in each position of the feature map.

\begin{figure*}[htb!]
	\centering
	    \includegraphics[width=0.8\columnwidth]{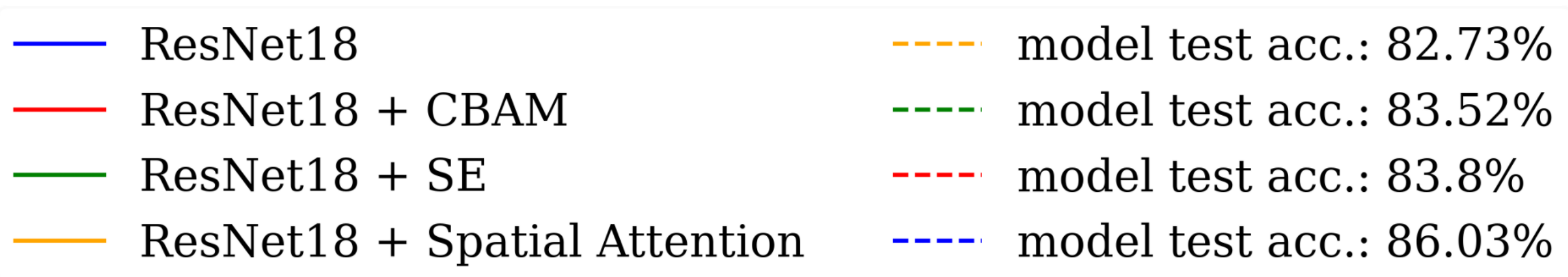}
	\vspace{-.3cm}
	
	\subfloat[LRP accuracy $p_l$\label{fig:attentionLRP1}]{
	    \includegraphics[width=0.95\columnwidth]{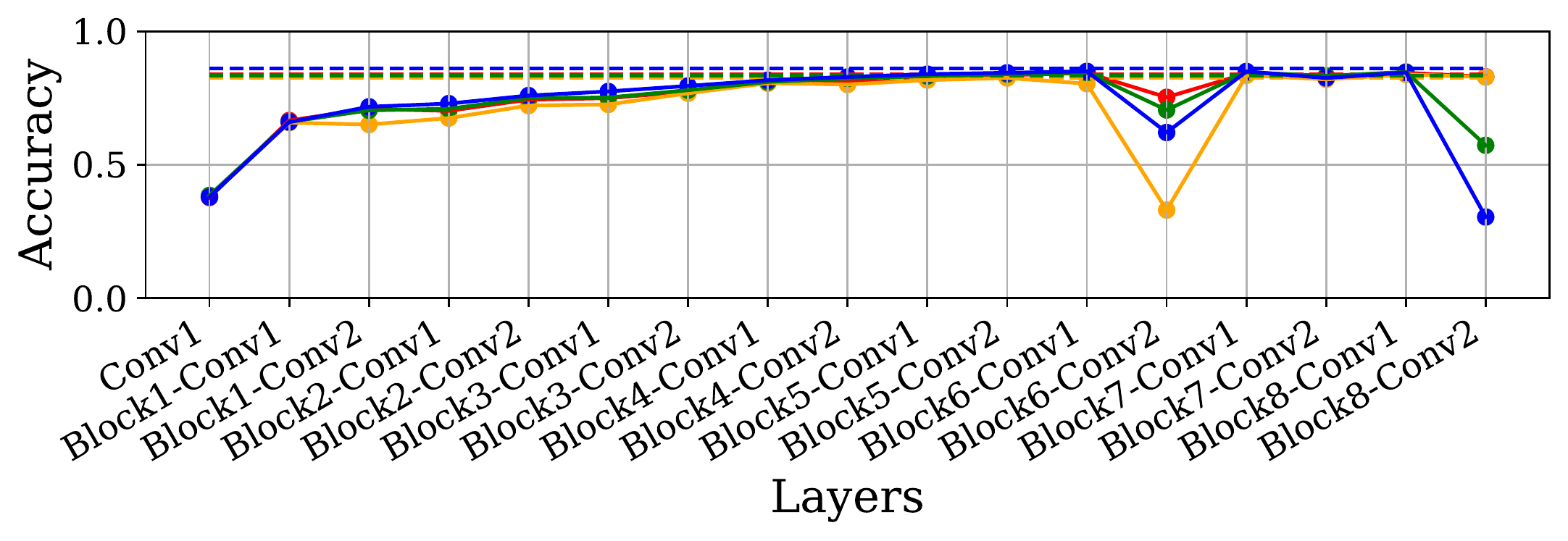}
	} \hspace{0.05\columnwidth}
		\subfloat[saturation values $s_l$\label{fig:attentionSat1}]{
	    \includegraphics[width=0.95\columnwidth]{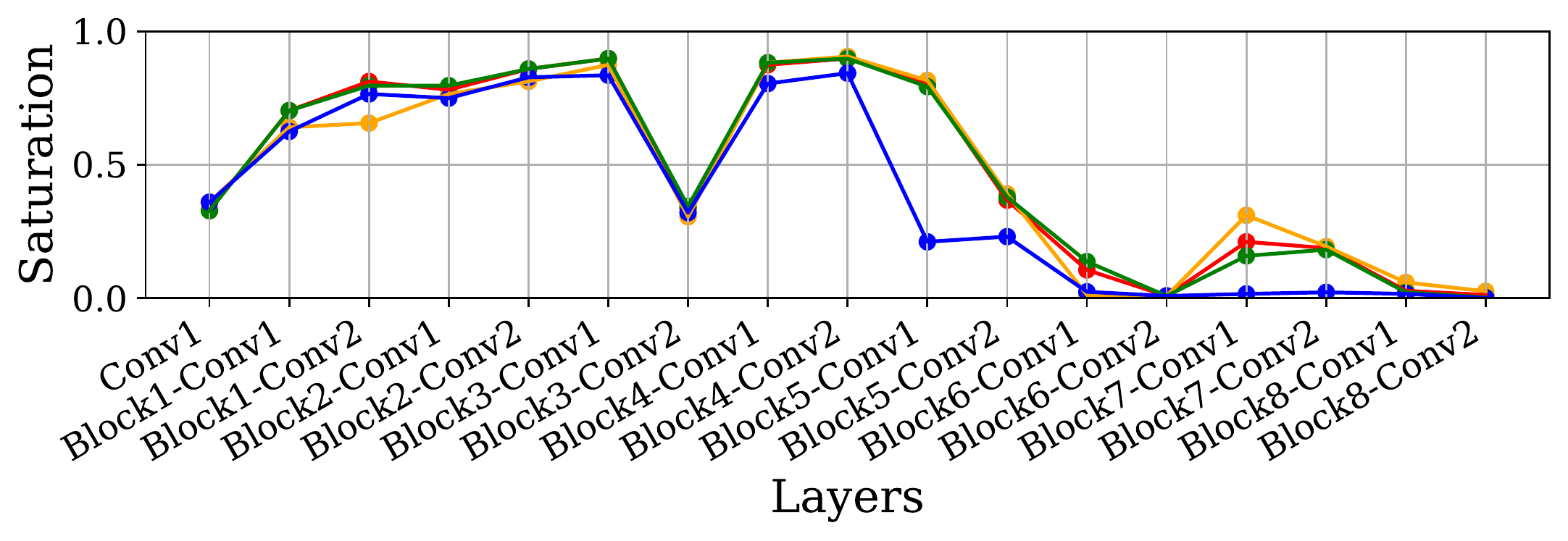}
	} \vspace{-.05cm}
	\subfloat[LRP accuracy $p_l$---with disabled skip connections \label{fig:attentionLRP2}]{
	    \includegraphics[width=0.95\columnwidth]{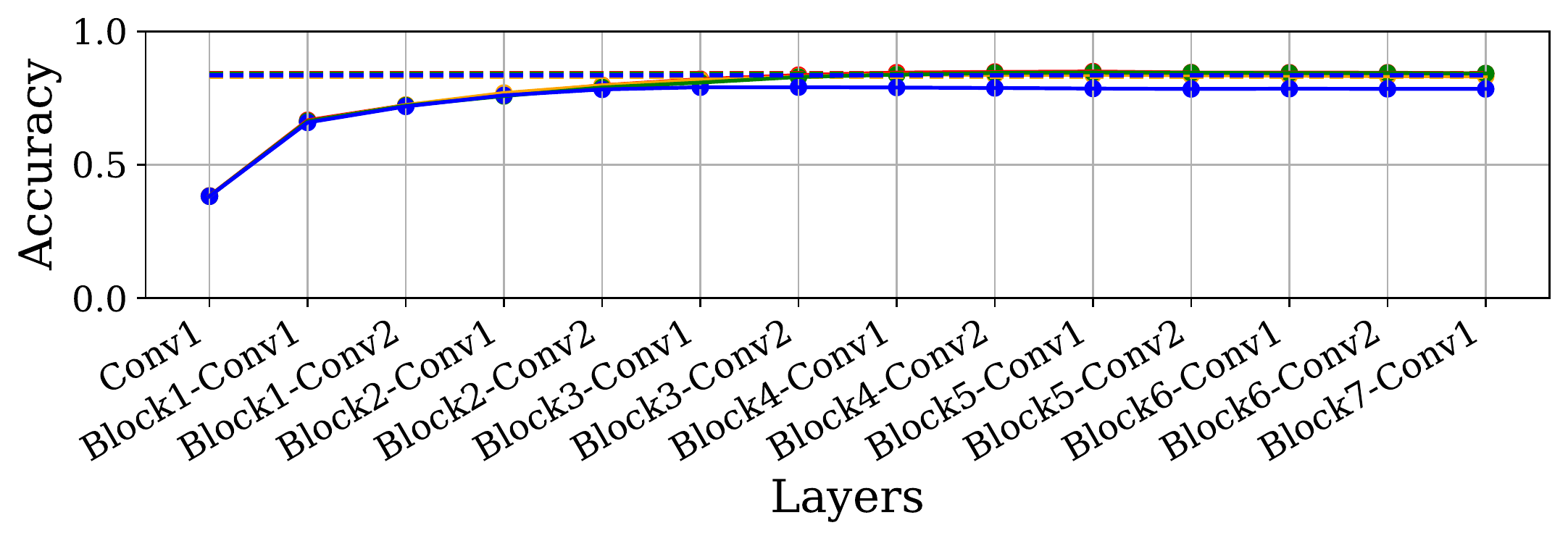}
	} \hspace{0.05\columnwidth}
	\subfloat[saturation values $s_l$---with disabled skip connections\label{fig:attentionSat2}]{
	    \includegraphics[width=0.95\columnwidth]{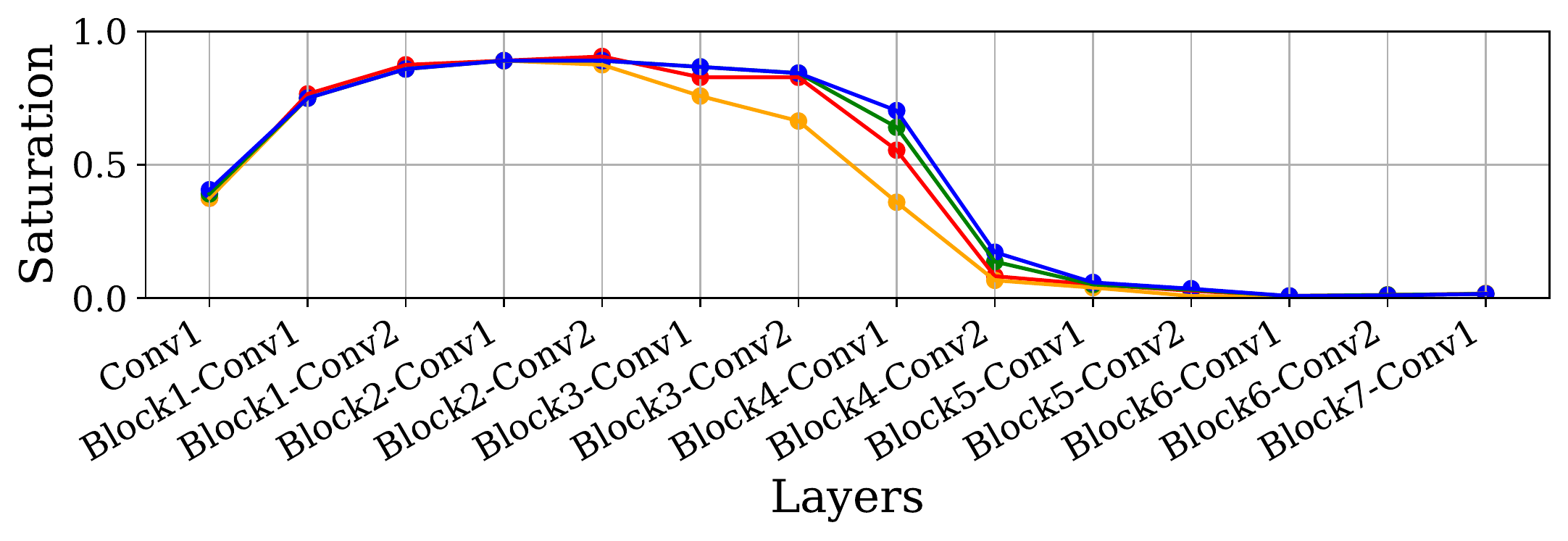}}
	\caption{The attention mechanisms Squeeze-and-Exitation modules (SE), spatial attention, and CBAM added to the ResNet 18 architecture with skip connectors \protect\subref{fig:attentionLRP1}, \protect\subref{fig:attentionSat1} and  without skip connectors \protect\subref{fig:attentionLRP2}, \protect\subref{fig:attentionSat2}. Attention mechanisms do neither change the LRP accuracy nor the saturation value. This indicates that attention mechanisms aid the feature extraction for the receptive field size present, but do not change the size of features extracted.}
	\label{fig:attention}
    \vspace{-0.15cm}
\end{figure*}

For testing this hypothesis, the attention mechanisms SE, 
spatial attention, and CBAM 
are added to a ResNet18 architecture. To ensure that there are no unexpected interactions between skip connections and attention mechanisms, we also consider ResNet18 architectures with attention mechanisms but with disabled skip connections.
Results for all experiments are shown in Fig.~\ref{fig:attention} (note that the additional layers added by the attention mechanism are omitted in that figure to make the sequence of values comparable to previous experiments).

The results support the hypothesis that there are no influences on the LRP and saturation values by any attention mechanisms. While the model accuracy varies depending on the choosen attention mechanism, LRP accuracy and saturation values do not change substantially.
Further, the plot implies that the attention mechanisms primarily change the way a layer extracts the features, while they apparently do not change the distribution of the overall inference.
Consequently, the border layer assumption is not influenced by any attention mechanism and can be used to predict the unproductive convolutional layers within the architecture.

\section{Implications on CNN Design}\label{sec:design}
Since the border layer can distinguish the unproductive sequences of convolutional layers, given only the architecture and the input resolution, the border layer can help to design and improve CNN architectures. 
A significant advantage of this border layer approach is, that both properties---network architecture and input image size---are known before starting training. Thus, our approach allows a very efficient and sustainable design process, since the architecture can be optimized without requiring comparative evaluation of trained models.

\begin{figure*}[ht!]
	\centering
	\subfloat{
	    \includegraphics[width=0.7\columnwidth]{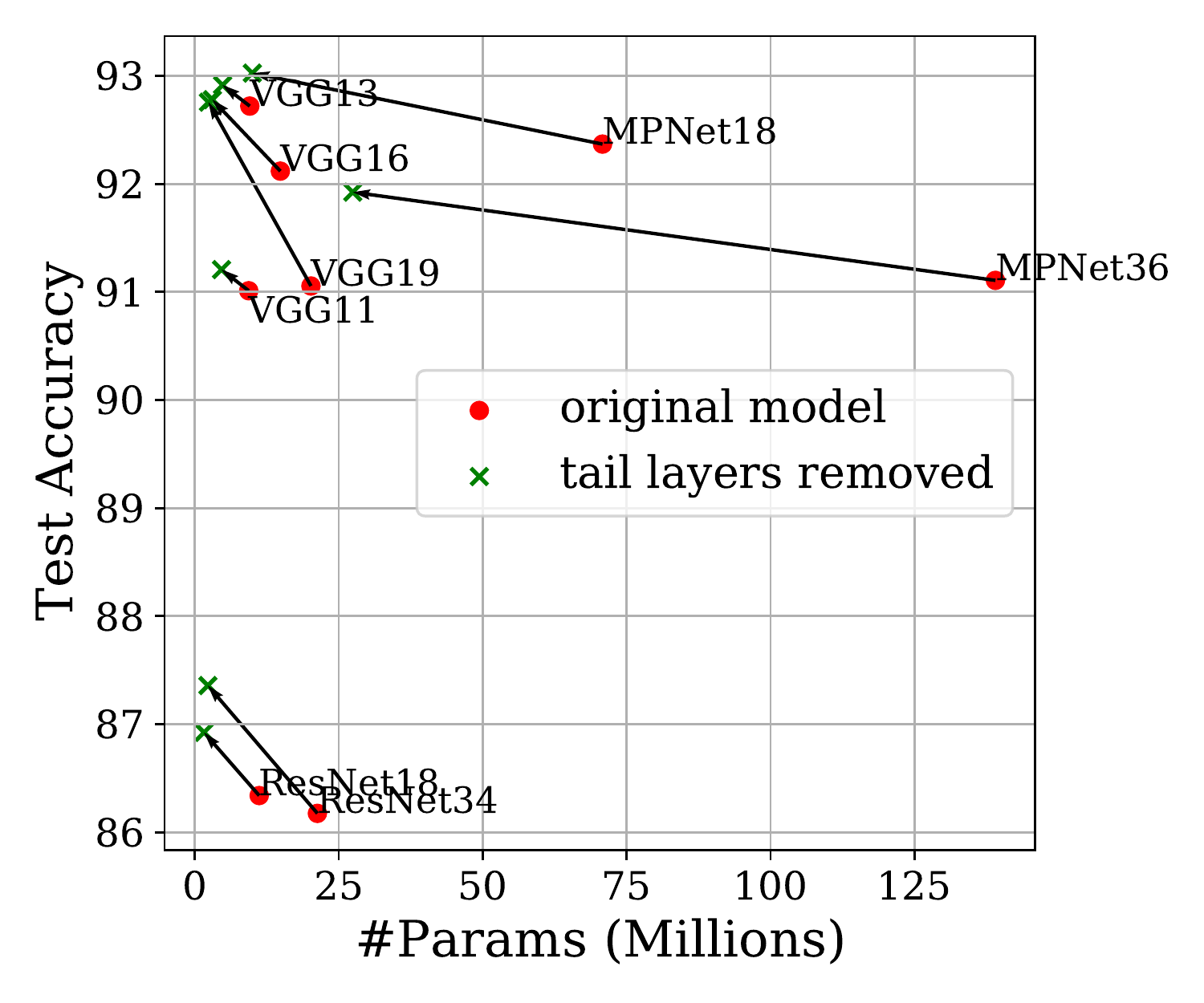}
	} \hspace{0.3\columnwidth}
	\subfloat{
	    \includegraphics[width=0.7\columnwidth]{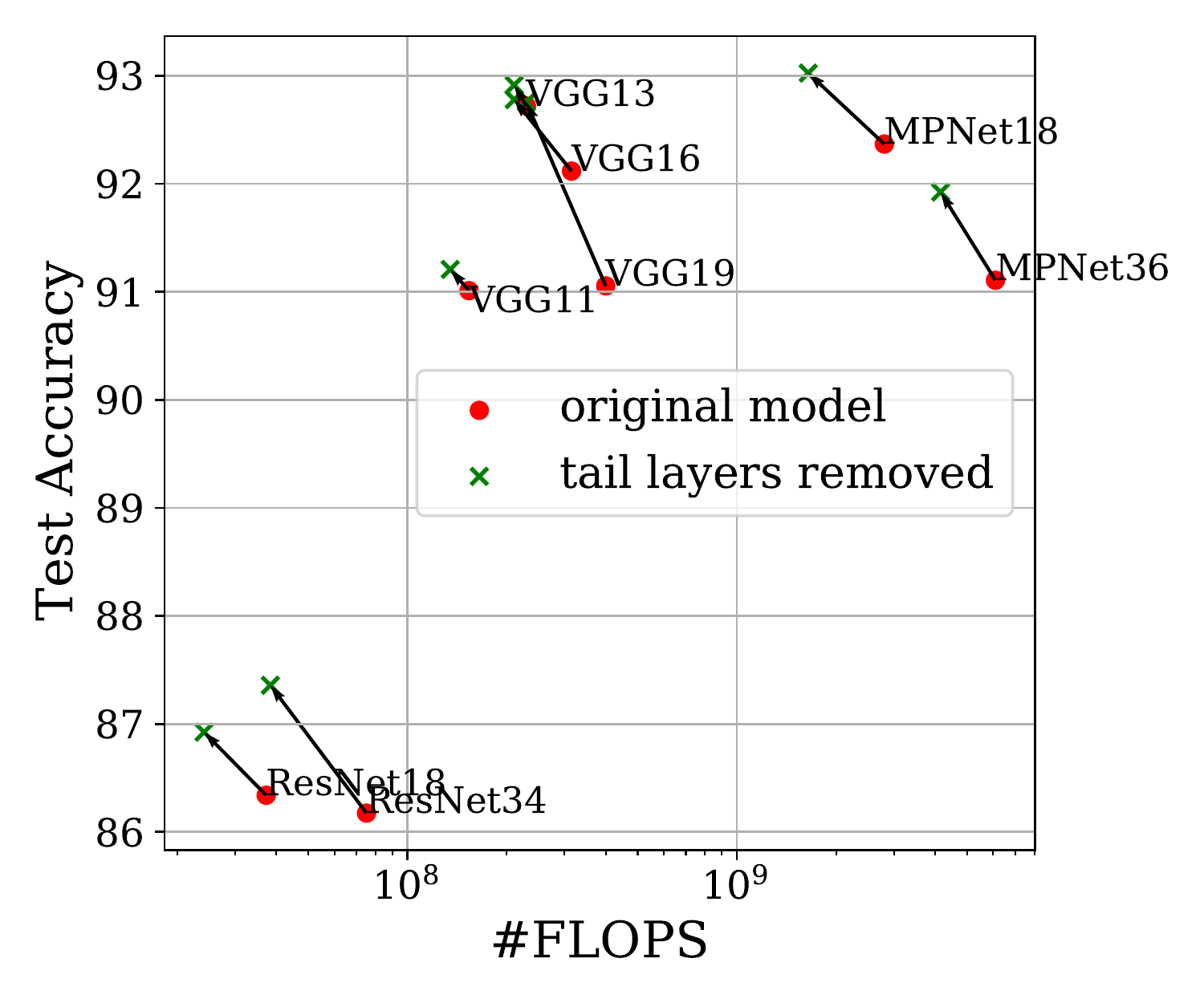}
	}
	\caption{A simple exemplary modification based on receptive field analysis. All layers with $r_{l-1, min} > i$ are replaced by a simple output-head, effectively removing the border layer and every layer after it. Removing these layers improves the efficiency by improving the performance as well as reducing the number of parameters and computations required.}
	\label{fig:efficiency}
	\vspace{-0.5cm}
\end{figure*}

The potential of the border layer approach is exemplified by optimizing existing architectures. On the Cifar10 dataset, the border layer is used to replace all unproductive layers together by a simple classifier layer, consisting of a global average pooling layer followed by a softmax layer.
We apply this optimization strategy on the VGG, ResNet and MPNet models used throughout this work.
Covering possible random fluctuations during training, all optimized models are trained ten times, and the average test accuracy, the total number of trainable parameters, and FLOPs required for a forward pass of a single image are computed. The resulting numbers are visualized in Fig.~\ref{fig:efficiency}.
It can be seen that the test accuracy of the optimized models improves in all tested scenarios while the removal of the unproductive layers reduced the number of trainable parameters. Thus, the FLOPs required for the computations are also reduced.

The demonstrated border layer technique, ensuring $r_{l-1} < i$, can be considered primitive and is not necessarily the most sophisticated or optimal way to optimize the CNN architecture for parameter and computational efficiency.
An alternative approach for optimizing the architecture, with a strong focus on test accuracy, is to influence the growth rate of the receptive field size. Influencing the growth rate can be achieved by e.g. adding, removing, and re-positioning pooling layers.
As an example, by removing the first two downsampling layers within the stem of ResNet18 and ResNet34, receptive field sizes in the entire network are reduced by a factor of 4. 
Doing so improves the Cifar10 test accuracy of ResNet34 from 82.76\% to 92.21\% and ResNet18`s from 84.61\% to 91.95\%.

By influencing the growth rate of the receptive field from within the network, an improvement in test accuracy and parameter efficiency is notable.
However, the removal of downsampling layers will at the same time increase the FLOPS per image, since the adequate size of the feature map is increased in every layer.
This design decision can be considered a trade-off in predictive quality over computational efficiency.
In the case of ResNet34, the computations per image increase from 0.76 GFLOPs to 1.16 GFLOPs and for ResNet18 from 0.04 GFLOPs to 0.56 GFLOPs.


In summary, based on the interaction between input resolution and receptive field size, the border layer can be used to predict unproductive layers.
Changing the network architecture to put the border layer as close to the output layer as possible is a suitable heuristic to optimize the CNN for computational efficiency, parameter efficiency and predictive performance, thereby making ideally all layers contribute qualitatively to the inference process.
Those architectural changes still are subject to a trade-off between predictive performance, parameter efficiency, and computational efficiency. 
This dilemma of optimization is the reason why this paper does provide a heuristic and not an optimization or pruning algorithm, since different application scenarios may require different compromises. 
Nevertheless, this heuristic allows the practitioner to make these design decisions with intent, allowing for an informed answer to the question "Should You Go Deeper".
\section{Conclusion}
While building ever deeper and more powerful neural architectures often is important for pushing the state-of-the-art, we think that also designing lightweight and efficient CNNs is crucial for opening deep learning to a broader range of applications with limited resources.
The results presented in this work allow users to detect and resolve inefficiencies in CNN architectures reliably without training the model, by the use of the border layer, i.e., by the receptive field size.
By exploring the properties of multipath architectures and architectures with skip connection and attention mechanisms, we have covered a broad range of common architectural components used in modern network designs.  
The achieved results are incorporated into our design heuristic based on the receptive field size, the LRP accuracy, and the saturation values.
This heuristic allows for degrees of freedom in the design process that directly enable the data scientist to make decisions in the optimization dilemma between computational power, predictive performance, or parameter efficiency.
Designing CNN architectures to be more parameter efficient could lead to lightweight, purpose-built models that need less training data and can be used, e.g., in embedded hardware platforms, for enabling an even broader field of potential applications. As shown in this work, by eliminating unproductive layers in the CNN architectures, the sustainability and viability of deep learning solutions will increase significantly.

\bibliographystyle{IEEEtran}
\bibliography{cleanBib.bib}

\end{document}